\documentclass[a4paper,11pt]{article}

\usepackage{amsfonts}
\usepackage{amsmath}
\usepackage{amsthm}
\usepackage[auth-sc,affil-sl]{authblk}
\usepackage{bm}
\usepackage{graphicx} 
\usepackage{caption}
\usepackage{subcaption}
\usepackage{multirow}
\usepackage{verbatim}
\usepackage{enumitem}
\usepackage{longtable}
\usepackage{float}
\usepackage{hyperref}
\hypersetup{
    colorlinks=true,
    linkcolor=blue,
    citecolor=blue,
    urlcolor=black
}
\usepackage{geometry}
\usepackage{underscore}
\usepackage{amssymb}
\usepackage{booktabs}
\usepackage{tabularx}
\usepackage{array}
\usepackage{makecell}
\usepackage{xcolor}
\usepackage{natbib}
\usepackage{siunitx}
\renewcommand
\baselinestretch{1.0} 
\usepackage{url}
\usepackage{natbib}
\usepackage{graphicx}
\usepackage{epstopdf} 
\graphicspath{{figures/}} 
\usepackage{lscape}
\usepackage[svgnames]{xcolor}
\usepackage{sansmath}
\usepackage{subcaption}
\usepackage{rotating}
\usepackage{tikz}
\usetikzlibrary{positioning, arrows.meta, shapes.geometric}
\usepackage{xspace}

\DeclareMathAlphabet{\mathpzc}{OT1}{pzc}{m}{it}

\newcommand \address[1]{\gdef \@address{#1}}
\makeatletter
\long\def\@footnotetext#1{\insert\footins{\def\baselinestretch{1.2}\footnotesize
\interlinepenalty\interfootnotelinepenalty
\splittopskip\footnotesep \splitmaxdepth \dp\strutbox
\floatingpenalty \@MM \hsize\columnwidth \@parboxrestore
\edef\@currentlabel{\csname
p@footnote\endcsname\@thefnmark}\@makefntext
{\rule{\z@}{\footnotesep}\ignorespaces #1\strut}}}

\long\def\symbolfootnote[#1]#2{\begingroup%
\def\thefootnote{\fnsymbol{footnote}}\footnote[#1]{#2}\endgroup}

\providecommand{\keywords}[1]{\textit{Keywords:} #1}

\def\maketitle{%
  \null
  \thispagestyle{empty}%
  \begin{center}\leavevmode
    \normalfont
    {\LARGE \bf \@title\par}%
    {\normalsize \@author\par}%
    \vskip 0.05 cm
    \vskip 0.05cm
    {\normalsize \@date\par}%
  \end{center}%
}

\makeatletter
\newcommand{\institute}[1]{\newcommand{\@institute}{#1}}

\renewcommand\textsl{\textcolor{blue}}

\begin{document}
\title{Interval Prediction of Annual Average Daily Traffic on Local Roads via Quantile Random Forest with High-Dimensional Spatial Data}
\author[]{Ying Yao$^*$}
\author[]{Daniel J. Graham}
\affil[]{Department of Civil Engineering, Imperial College London, London, SW7 2AZ, UK.\\ $^*$ Corresponding Author: \texttt{ying.yao23@imperial.ac.uk}} 
%

\date{}

\maketitle

\begin{abstract}
Accurate annual average daily traffic (AADT) data are vital for transport planning and infrastructure management. However, automatic traffic detectors across national road networks often provide incomplete coverage, leading to underrepresentation of minor roads. While recent machine learning advances have improved AADT estimation at unmeasured locations, most models produce only point predictions and overlook estimation uncertainty. This study addresses that gap by introducing an interval prediction approach that explicitly quantifies predictive uncertainty. We integrate a Quantile Random Forest model with Principal Component Analysis to generate AADT prediction intervals, providing plausible traffic ranges bounded by estimated minima and maxima. Using data from over 2,000 minor roads in England and Wales, and evaluated with specialized interval metrics, the proposed method achieves an interval coverage probability of 88.22\%, a normalized average width of 0.23, and a Winkler Score of 7,468.47. By combining machine learning with spatial and high-dimensional analysis, this framework enhances both the accuracy and interpretability of AADT estimation, supporting more robust and informed transport planning.

\keywords{Annual Average Daily Traffic; Interval Prediction; Quantile Random Forest; \quad Principal Component Analysis; High-Dimensional Spatial Data}.

\end{abstract}

\section{Introduction}
Average Annual Daily Traffic (AADT), measured as the mean number of vehicles traversing a road link per day, is a fundamental metric in transportation analysis. It supports data-driven decisions across engineering, planning, economics, and environmental management. AADT informs both long-term strategies such as infrastructure investment, capacity planning, and network design, and short-term operations including signal optimization, congestion mitigation, safety management, emissions control, and maintenance scheduling.

In practice, AADT is derived from data collected by Automatic Traffic Counters (ATCs). While major roads (freeways and Class A roads) are well instrumented, minor roads (Class B and C) typically receive limited monitoring due to lower traffic volumes. In England and Wales, for instance, minor roads represent around 87\% of total road length but only 13\% are equipped with detectors. This imbalance creates substantial data gaps and limits the comprehensiveness of national AADT datasets.

To address missing data, AADT values are often estimated using statistical or machine learning models that exploit historical traffic, link characteristics, demographics, and environmental variables. However, most existing approaches yield only point estimates, that is, single expected values that fail to reflect underlying uncertainty or variability in traffic conditions. This limitation can obscure atypical patterns and lead to overconfident planning decisions.

This study introduces an interval prediction framework that explicitly quantifies predictive uncertainty, providing a plausible range of AADT values bounded by estimated minima and maxima. Interval-based estimation offers a more informative foundation for transportation decision-making by highlighting variability that may influence congestion management, capacity utilization, safety interventions, and risk assessment.

Our approach employs a Quantile Random Forest (QRF) model integrated with Principal Component Analysis (PCA) for dimensionality reduction. The model is trained on a high-dimensional feature space of 888 variables spanning geographic attributes, accessibility measures, infrastructure characteristics, spatial lags, density indicators, and multi-scale buffer statistics (500–3,000 m). Bayesian optimization and stochastic search are used for hyperparameter tuning, while interval performance is evaluated primarily using the Winkler Score (WS). Through this procedure, we construct a QRF model that demonstrates strong generalization ability and delivers accurate interval forecasts for unseen data \citep{raykov2014population, enwright2019modeling}. 

Using data from over 2,000 minor roads in England and Wales for 2021, we demonstrate that the proposed model effectively recovers AADT distributions informed by rich spatial and contextual variables. The study contributes to the literature by (i) applying QRF to interval-based AADT estimation with a high-dimensional feature set, (ii) analysing the trade-off between interval coverage and width, and (iii) showcasing the value of uncertainty quantification in transportation planning.

The remainder of this paper is organized as follows: Section 2 reviews related work on AADT prediction and machine learning. Section 3 outlines the proposed methodology, including feature construction, PCA, QRF modelling, and performance evaluation. Section 4 presents the model results and interval assessment using the Risk Assessment Index (RAI) and Winkler Score. Section 5 illustrates two applications of interval estimation in traffic management and planning. Section 6 discusses findings and connections to existing studies, followed by conclusions in Section 7.

\section{Literature Review}

\subsection{AADT Estimation Models}
The concept of Annual Average Daily Traffic (AADT), introduced in the 1980s, represents the mean 24-hour traffic volume (both directions) over a year. It is typically computed by dividing the total annual vehicle count at a given location by 365 days (or 366 in leap years) \citep{pricker1987traffic}.

AADT serves as a core indicator in transportation research and practice. It supports analyses of highway accident frequencies \citep{tortum2015artificial}, network modelling, infrastructure planning, and congestion management \citep{leduc2008road}. As a key input for calculating vehicle kilometres travelled (VKT), AADT informs maintenance and investment strategies \citep{leduc2008road, wang2013estimating}, while also contributing to roadway design, operational management \citep{fu2018estimating}, and road safety studies \citep{wang2013estimating}. Recent work has expanded its use to modelling road transport activity and emissions, underpinning greenhouse gas inventories, IEA-compliant reporting, traffic simulation, and long-term planning \citep{ganji2020methodology, fu2018estimating}.

AADT data are primarily collected through permanent automatic traffic recorders, supplemented by short-term counts adjusted with correction factors \citep{leduc2008road}. However, complete spatial coverage is impractical—particularly for minor roads—necessitating predictive estimation at unmonitored sites \citep{baffoe2022estimation}. Point-based models address this need by inferring traffic volumes from nearby stations, valued for their simplicity and scalability \citep{wang2023comparison}.

Traditional modelling approaches include the four-step framework—trip generation, trip distribution, modal split, and traffic assignment \citep{apronti2018four}. Although useful for regional-scale forecasting, this method requires large datasets and struggles with nonlinear interactions at the local level. Classical regression techniques have also been applied to link traffic volumes with roadway, land-use, and socioeconomic attributes (e.g., \citep{sfyridis2020annual, chen2019estimating, zhao2001contributing, wang2009forecasting}), but they often suffer from multicollinearity in high-dimensional settings \citep{jayasinghe2019novel}.

Spatial statistical methods have been used to improve AADT estimation by capturing spatial dependence and interpolation effects. Examples include spatial regression and kriging models that enhance accuracy for off-network facilities \citep{eom2006improving}, as well as hybrid approaches combining time-series and spatial analysis to integrate historical, environmental, and infrastructural data \citep{pun2019multiple, chen2019estimating}. Incorporating socioeconomic and geographic correlations through clustering and regression has also improved predictive performance, increasing $R^2$ from 0.46 to 0.75 in some cases \citep{seaver2000estimation}.

Machine learning has emerged as a powerful alternative for AADT prediction. Commonly used algorithms include Support Vector Regression (SVR) \citep{khan2018development, sun2019estimating}, Random Forests (RF) \citep{sfyridis2020annual, georganos2019geographical, fouedjio2024locally}, and K-prototypes clustering \citep{sfyridis2020annual}. Model performance is typically evaluated using $R^2$, Root Mean Square Error (RMSE), and Mean Absolute Percentage Error (MAPE) \citep{baffoe2022estimation}. To mitigate class imbalance, the Synthetic Minority Oversampling Technique (SMOTE) is often employed \citep{Han2025}.

\subsection{Interval Prediction Method}

In light of these limitations of point prediction models, we now turn to interval prediction approaches, which explicitly account for uncertainty in traffic flow estimation. In this brief review, we outline several widely used interval prediction methods, including the Quantile Random Forest (QRF) algorithm that forms the basis of our study. For clarity, we group existing methods into three main categories, which are summarized below and in Table~\ref{LRS} in the appendix.

\subsubsection{Statistical interval prediction.}
    
    Recent studies illustrate the diversity of statistical approaches to interval prediction across different domains. In the context of meta-analysis, prediction intervals (PIs) have been emphasized as a complement to conventional confidence intervals (CIs). A typical CI for the overall mean effect can be written as

   \begin{equation}
    CI = \hat{\mu} \,\pm\, z_{\alpha/2} \, SE\!\left(\hat{\mu}\right),
    \label{eq:CI}
    \end{equation}

    where $\hat{\mu}$ is the estimated mean effect and $SE(\hat{\mu})$ is its standard error. By contrast, the PI aims to quantify the range of future effects:
    
    \begin{equation}
\label{eq:PI}
PI = \hat{\mu} \,\pm\, t_{\alpha/2,\,k-2}\, 
\sqrt{\tau^{2} + SE^{2}\!\left(\hat{\mu}\right)},
\end{equation}
    
    where $\tau^2$ captures between-study variance. Comparative studies based on random-effects meta-analysis using the Hartung–Knapp–Sidik–Jonkman method (REMA, HKSJ, EB, PI) demonstrate that PIs provide more informative reflections of heterogeneity and the expected variability in future studies \citep{Botella2024}.

    Other work has proposed unified pivotal frameworks, such as skew-normal prediction quantile intervals (SN-PQ-Intervals), which extend standard methods by directly estimating conditional quantiles. In such cases, the $\tau$-quantile of outcome $Y$ given predictors $X$ can be expressed as
    
    \begin{equation}
    Q_Y(\tau \mid X) = X^\top \beta_\tau,
    \end{equation}
    
    with intervals constructed from lower and upper quantile functions. Simulation results confirm the robustness of these methods under skewed distributions \citep{Qi2022}.
    
    Systematic reviews also reveal practical challenges. For example, analysis of Cochrane reviews showed that a large share of significant findings had PIs crossing the null value, and in some cases even suggesting opposite effects. This highlights the necessity of routinely reporting prediction intervals in evidence synthesis \citep{IntHout2016}.

    Beyond biomedical applications, interval prediction has been extended to financial markets. Grey system-based formulations, such as the grey interval prediction approach, incorporate dynamic adjustment of interval width to improve reliability in short-term stock index forecasting \citep{Xie2014}.
    
    Finally, alternative transformations have been proposed to improve interval precision across different distributions. The Box-Cox exponential transformation (Box–Cox ET), for instance, applies
    
    \begin{equation}
    Y^{(\lambda)} = \frac{Y^{\lambda} - 1}{\lambda}, \quad \lambda \neq 0,
    \end{equation}
    
    to stabilize variance before constructing intervals, yielding more accurate and consistent coverage for distributions including normal, exponential, Weibull, and lognormal cases. Simulation studies confirm the effectiveness of such transformation-based methods \citep{Yu2009}.

    \subsubsection{Machine learning interval prediction.}
    
    Machine learning–based interval prediction methods have developed rapidly, particularly with the introduction of ensemble and deep learning architectures. A key example is the application of quantile regression random forests (QRRF) combined with deep neural network structures such as VGGNet, ResNet, Inception, and long short-term memory networks (LSTM). This framework was tested on the Singapore electricity market and demonstrated accurate and reliable probabilistic forecasts by exploiting both tree-based ensembles and deep feature extraction \citep{Dang2022}.

    Another important line of research focuses on adapting random forest (RF) algorithms to large-scale datasets. Parallel implementations, online RF learning, and subsampling strategies have been developed to improve computational efficiency and robustness to noise. Applications to datasets with millions of records, such as airline scheduling, show that these extensions maintain high predictive accuracy even in noisy, high-dimensional environments \citep{Genuer2017}.

    A foundational contribution was the introduction of quantile regression forests (QRF), which extend the RF framework from conditional mean estimation to conditional quantile estimation. Formally, the conditional quantile of a response variable $Y$ given predictors $X$ at quantile level $\tau$ is estimated as

    \begin{equation}
    \hat{Q}_Y(\tau \mid X = x) = \sum_{i=1}^n w_i(x) y_i,
    \end{equation}
    
    where $w_i(x)$ are data-dependent weights derived from the frequency with which training samples fall into the same leaf nodes as $x$ across the ensemble of trees. Prediction intervals are then constructed as
    
    \begin{equation}
\label{eq:PI_quantile}
PI_{\alpha}(x) = 
\Bigl[\, 
\hat{Q}_{Y}\!\left(\tfrac{\alpha}{2}\,\bigm|\, X = x \right),\;
\hat{Q}_{Y}\!\left(1 - \tfrac{\alpha}{2}\,\bigm|\, X = x \right) 
\,\Bigr].
\end{equation}

    Empirical studies using housing, environmental, and economic datasets confirmed that QRF produces well-calibrated prediction intervals, offering robustness against noise and irregular distributions \citep{Meinshausen2006}.

    \subsubsection{Hybrid statistical–machine learning interval prediction.}
    
   Hybrid methods integrate the explanatory power of statistical modeling with the adaptability of machine learning, which enhances the reliability and robustness of prediction intervals. One recent study applied deep neural networks (DNN) with a hybrid loss ensemble on benchmark regression datasets such as housing and wine quality. The loss function balances point prediction accuracy with interval reliability:
    
    \begin{equation}
\label{eq:loss}
\mathcal{L} 
= \operatorname{MSE}(y, \hat{y})
+ \lambda \cdot \operatorname{MPIW}
+ \eta \cdot \operatorname{PICP\_penalty},
\end{equation}

    where MPIW denotes the mean prediction interval width and the penalty term ensures that the prediction interval coverage probability (PICP) remains above the desired confidence level. This design achieved 95\% coverage with significantly narrower intervals, handling both aleatory and epistemic uncertainties \citep{lai2022exploring}.  
    
    In financial and energy forecasting, decomposition and distributional modeling have been combined with optimization techniques. For example, a study of EU carbon emission trading prices employed time-varying filtering with empirical mode decomposition (TVF-EMD) and IS-MODA optimization, fitting a lognormal distribution for point and interval prediction:
    
    \begin{equation}
\label{eq:lognormal_PI}
\begin{split}
&\ln(P_t) \sim \mathcal{N}(\mu, \sigma^{2}), \\
&PI = \Bigl[\,
\exp\!\left(\mu - z_{\alpha/2}\sigma\right),\;
\exp\!\left(\mu + z_{\alpha/2}\sigma\right)
\,\Bigr].
\end{split}
\end{equation}

    This hybrid structure delivered high accuracy, with mean absolute percentage errors (MAPE) between 0.8\% and 1.1\%~\citep{niu2022combined}.  
    
    Extensions of quantile regression random forests (QRRF) have also been proposed. By integrating whale optimization algorithms (WOA), discrete wavelet transform (DWT), and risk assessment indices (RAI), the model minimizes prediction interval width subject to coverage constraints:
    
    \begin{equation}
    \min \; \text{MPIW}, \quad \text{s.t.} \quad \text{PICP} \geq 1-\alpha,
    \end{equation}
    
    while conditional quantiles are estimated as
    
    \begin{equation}
    \hat{Q}_Y(\tau \mid X) = \sum_{i=1}^n w_i(X) y_i,
    \end{equation}
    
    with weights $w_i(X)$ derived from the RF structure. This framework provided accurate probabilistic load forecasting and incorporated risk assessment directly into interval estimates~\citep{aprillia2021statistical}.  
    
    Financial markets also benefit from hybrid semi-parametric methods. A Generalized Autoregressive Conditional Heteroskedasticity–Quantile Regression Random Forest (GARCH–QRRF) framework links variance dynamics with quantile estimation:
    
    \begin{equation}
    \sigma_t^2 = \omega + \alpha \epsilon_{t-1}^2 + \beta \sigma_{t-1}^2,
    \end{equation}
    
    and
    
    \begin{equation}
    VaR_{\tau,t} = \mu_t + \sigma_t \hat{Q}_Z(\tau),
    \end{equation}
    
    where $\hat{Q}_Z(\tau)$ is the quantile of standardized residuals. Empirical applications to stock indices show that this method outperforms parametric benchmarks, especially under heavy-tailed $t$ and GED distributions~\citep{jiang2017semi}.  
    
    Finally, hybrid methods have been applied in transport studies. An artificial neural network (ANN) combined with discriminant analysis was used to classify seasonal traffic volume variations across 86 stations. The discriminant function is expressed as
    
    \begin{equation}
    D_k(x) = x^\top \Sigma^{-1} \mu_k - \tfrac{1}{2}\mu_k^\top \Sigma^{-1} \mu_k + \ln \pi_k,
    \end{equation}
    
    where $\mu_k$ is the class mean and $\pi_k$ is the prior probability. The hybrid model achieved 100\% classification accuracy, identifying spatial links, regional effects, and road function as the most important explanatory factors~\citep{Splawinska2018}.

Two widely adopted metrics are typically employed to assess the performance of interval prediction models. The Risk Assessment Index (RAI) integrates the Normalized Average Width (NAW) with the Prediction Interval Coverage Probability (PICP), offering a comprehensive measure of model effectiveness in complex data settings. The Winkler Score (WS) is another important indicator that evaluates the accuracy of prediction intervals by penalizing those that fail to capture the true value while rewarding narrower intervals when the true value is included \citep{aprillia2021statistical}. Optimal parameter settings can be identified by calculating and comparing RAI values across different parameter combinations, which in turn improves both predictive accuracy and the generalization capacity of the model. In this study, model performance is assessed using RAI and WS \citep{aprillia2021statistical}.

\section{Methods}
The proposed methodology consists of five key stages: data preparation, dimensionality reduction via PCA, model development, hyperparameter optimization, and performance evaluation using interval-specific metrics. During the data preparation stage, cleaning procedures are applied to address missing values and to refine, filter, and normalize the dataset. PCA is then employed to reduce the dimensionality of the feature space while preserving essential information. Hyperparameter optimization is carried out to determine the best parameter configuration for model training. The optimized parameters together with the transformed features from PCA are subsequently used to construct the Quantile Random Forest (QRF) model for AADT prediction. Finally, the predictive accuracy and robustness of the model are assessed using diagnostic measures tailored for interval prediction. In addition, the 50\% quantile prediction interval is adopted as the representative point prediction output, and corresponding point prediction metrics are also employed for model evaluation.

\subsection{Data Preparation}
AADT data for the year 2021, covering more than 19,000 locations across England and Wales (EW), were obtained from the UK Department for Transport (DfT) \citep{dft2021road}. The dataset distinguishes between major and minor roads, where major roads consist of A roads and motorways, while minor roads include B roads, C roads, and unclassified categories. Although minor roads represent a substantial share of the UK road network, the available data indicate that only 23\% of recorded AADT corresponds to them. This study therefore develops a predictive model for AADT with a particular emphasis on minor roads.

In developing our predictive model, we utilise domain knowledge to derive zone level traffic potential predictors, and take a data driven approach over a high dimensional covariate vector to account for context and link specific factors. We now discuss each element in turn.

\subsection{Representing the traffic generation potential of zones}

Transportation engineering theory suggests that the traffic generation potential of zones can be adequately modeled via simple accessibility metrics (e.g., \citep{OrtuzarWillumsen2011}). Thus, as aggregate zone level predictors for traffic volumes we calculate two accessibility metrics based on gravity and negative exponential forms.      

\subsubsection{Gravity based accessibility metric}

The gravity based metric models the potential traffic flow between zones $i$ and $j$ using
\begin{equation}
t_{ij}=\frac{m_i \cdot m_j}{{d_{ij}^\alpha}},
\end{equation}
where $t_{ij}$ is traffic flow, $m_j$ is a measure of mass at zone $j$ (e.g. population or employment), and $d_{ij}$ is the Euclidean distance between zones $i$ and $j$, typically calculated by applying the Pythagorean theorem to the $x$ and $y$ coordinates of the zone centroids. The parameter $\alpha$ controls the strength of impedance over distance, that is the decay in volumes of interactions over space. We will assume that $\alpha$ is positive, and therefore, that traffic flow is a monotonically decreasing function of distance. 

To represent the total traffic generation potential of each zone $i$, which we denote $\rho_i$, we can simply sum over the traffic generated in all other zones $j$, giving 
\begin{equation}\label{rg1}    
\rho^G_i = \frac{1}{n} m_i \sum_{j=1}^n \frac{m_j}{d_{ij}^\alpha}.
\end{equation}
Note that when $i=j$ the Pythagorean theorem would calculate $d_{ii}$ at a single point, and thus give a Euclidean distance of zero, which is of course not appropriate as zones have internal mass and distance that we must represent. Instead we measure intra-zonal distances using the formula
\begin{equation}
d_{ii} = \sqrt{\frac{A_i}{\pi}} .
\end{equation}

where $A_i$ is the area of the zone.

\subsubsection{Negative exponential accessibility metric}
 
The negative exponential accessibility metric represents the total traffic generation potential of each zone using
\begin{equation}\label{re}     
\rho^E_i =  \sum_{j=1}^n m_j\cdot \exp(-\alpha d_{ij}),
\end{equation}
where again we assume that $\alpha$ is positive. Note that in this formulation we can allow $d_{ii}$ to take the value zero, since in this case the traffic generation potential of zone $i$ is then simply directly proportional to its mass $m_i$, which seems a reasonable proposition. 

\subsection{High dimensional feature space construction}

\subsubsection{Feature categorisation and selection}
The factors influencing AADT are incorporated as independent variables in the predictive models, together with the zone-level traffic potential indicators described previously. We differentiate between features defined at the point location and those derived from the service area (or zone). Point-based features include spatial lags of AADT, locational attributes (e.g., rural or urban classification, presence within built-up areas, proximity to functional urban areas and major towns), road characteristics, and traffic-related factors such as proximity to ports and airports. Service-area features consist of roadway measures (e.g., distance to motorway entrances and exits), socio-demographic variables (including number of businesses, household income, employment, population, and housing), and transport indicators (such as car ownership and access to public transport). By systematically categorizing and quantifying these variables, the models can more effectively capture and predict traffic flow patterns. After feature categorization and selection, the resulting dataset contains 19,616 observations and 910 variables.

\begin{figure}[!htbp]
    \centering 
    \resizebox{3.5in}{!}{%
\begin{tikzpicture}[
    node distance = 1.2cm,
    process/.style = {rectangle,rounded corners=2mm, draw, fill=orange!20, text width=3.5cm, 
                     minimum height=1.2cm, align=center, font=\small\bfseries},
    description/.style = {rectangle, draw, fill=blue!15, text width=7cm, 
                         minimum height=1.2cm, align=center, font=\small},
    arrow/.style = {->, >=stealth, thick, draw=blue!60!black},
    line/.style = {-, thick, draw=black}
]

\node[process] (step1) {Prepare \textbf{features} and \textbf{objective values}};
\node[process, below=of step1] (step2) {Filter out \textbf{minor road} data};
\node[process, below=of step2] (step3) {Calculate the percentage of non-missing values in each feature column};
\node[process, below=of step3] (step4) {Remove feature columns with greater than \textbf{25\% missing values}};
\node[process, below=of step4] (step5) {Remove rows containing any missing values};

\node[description, right=1cm of step1] (desc1) {y is AADT, all other factors form feature matrix X.};
\node[description, right=1cm of step2] (desc2) {Reduction in the number of data rows from 19,616 to 4,036.};
\node[description, right=1cm of step3] (desc3) {Calculate the proportion of missing data in each feature column, feature columns with more missing data will not be considered.};
\node[description, right=1cm of step4] (desc4) {Reduction in the number of data columns from 910 to 888.};
\node[description, right=1cm of step5] (desc5) {Reduction in the number of data rows from 4,036 to 2,247.};

\draw[arrow, line width=6pt] (step1) -- (step2);
\draw[arrow, line width=6pt] (step2) -- (step3);
\draw[arrow, line width=6pt] (step3) -- (step4);
\draw[arrow, line width=6pt] (step4) -- (step5);

\draw[line] (step1.east) -- (desc1.west);
\draw[line] (step2.east) -- (desc2.west);
\draw[line] (step3.east) -- (desc3.west);
\draw[line] (step4.east) -- (desc4.west);
\draw[line] (step5.east) -- (desc5.west);

\end{tikzpicture}
}
    \caption{Flowchart of data cleaning.}
    \label{fig:2}
\end{figure}
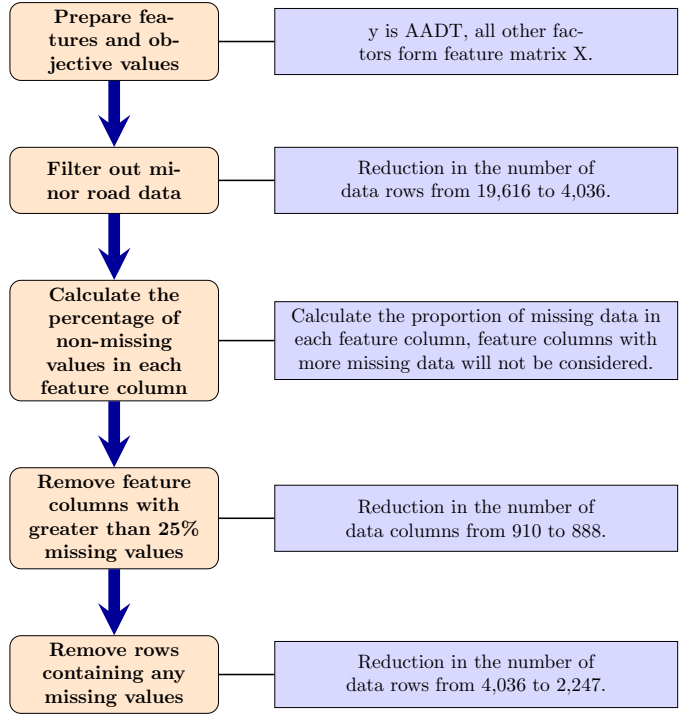

The initial filtering step extracted 4,036 records for minor roads from the original dataset of 19,616 entries. Given the considerable influence of missing values on model reliability, a thorough data cleaning procedure was implemented, as illustrated in Figure~\ref{fig:2}. Following this process, the final feature matrix comprised 2,247 observations and 888 variables.

The AADT values in the filtered dataset span a wide range, with most observations clustered at the lower end, resulting in a skewed distribution rather than a symmetric normal one. Applying a logarithmic transformation helps to reduce this skewness and approximate normality. In addition, it contributes to variance stabilization and enhances the accuracy of statistical inference \citep{james2023introduction}. For these reasons, a logarithmic transformation was applied to the AADT data.

\subsubsection{Principal Component Analysis}

Principal Component Analysis (PCA) reduces a potentially correlated set of variables into a smaller number of linearly independent components through orthogonal transformation. The resulting principal components capture the maximum variance within the data, offering an efficient and interpretable representation well suited for structured, high-dimensional datasets such as AADT features. PCA was preferred over alternatives like t-SNE or autoencoders for its computational efficiency, transparency, and robustness to correlated predictors.

Formally, given a data matrix $X \in \mathbb{R}^{n \times k}$ with rows $\{\mathbf{x}_1^T, \ldots, \mathbf{x}_n^T\}$, where $\mathbf{x}_i \in \mathbb{R}^k$, each vector can be expressed as a linear combination of an orthonormal basis $\{\mathbf{b}_1, \ldots, \mathbf{b}_k\}$:
\begin{equation}
  \label{eq::lincomb}
  \mathbf{x}_i = a_{i1}\mathbf{b}_1 + \cdots + a_{ik}\mathbf{b}_k,
\end{equation}
where $a_{ij}$ are the coordinates of $\mathbf{x}_i$ with respect to the basis $B$. PCA seeks an $r$-dimensional subspace $U \subseteq \mathbb{R}^k$ with orthonormal basis $\{\tilde{\mathbf{b}}_1, \ldots, \tilde{\mathbf{b}}_r\}$ such that
\begin{equation}
  \label{eq::PCAnewlincomb}
  \tilde{\mathbf{x}}_i = \sum_{j=1}^r \tilde{a}_{ij}\tilde{\mathbf{b}}_j
\end{equation}
provides the best low-rank approximation of $\mathbf{x}_i$. The basis vectors $\tilde{\mathbf{b}}_j$—the \textit{principal components}—are chosen sequentially to maximize captured variance.

PCA is particularly effective for this study’s dataset, which contains numerous interdependent geographic, demographic, and infrastructure-related variables. By transforming correlated features into orthogonal components, PCA enhances computational tractability and reduces redundancy without substantial information loss.

Previous studies have applied PCA to extract dominant features, identify latent structures, and simplify model development \citep{abdi2010principal}. Following this approach, 884 cleaned variables were grouped into 50 feature categories based on type and spatial radius (500–3,200 m), and PCA was applied to each group independently.

\begin{figure}[!t]
\centering
\captionsetup[subfigure]{labelformat=default}  
\subfloat[Group\_accessibility]{%
    \includegraphics[width=0.48\linewidth]{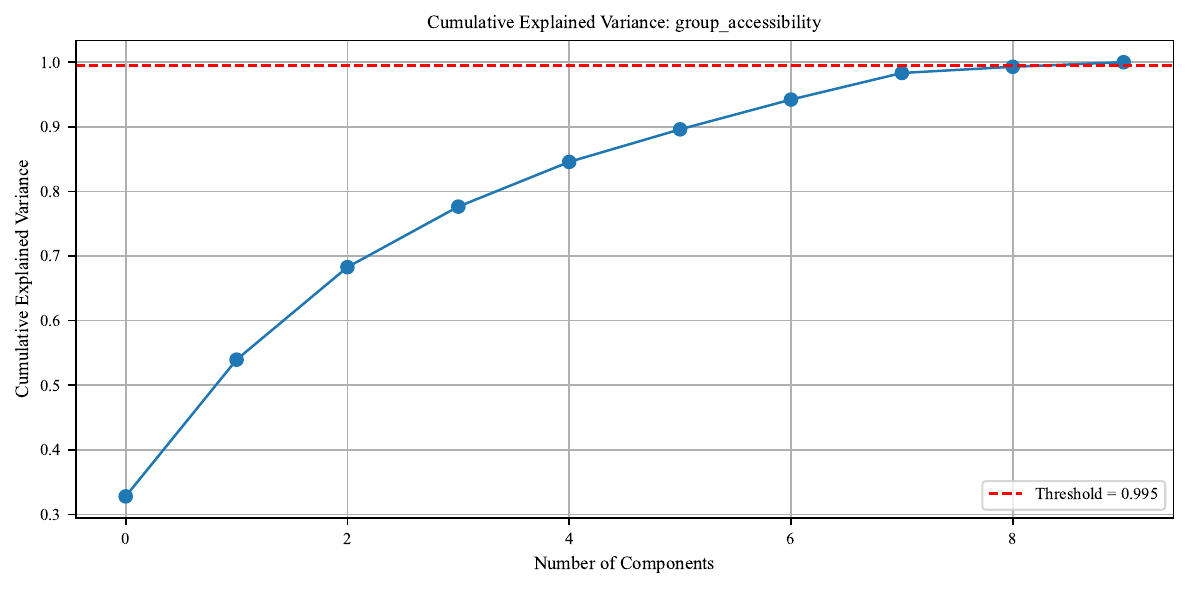}%
    \label{fig:3a}}
\hfill
\subfloat[Group\_BCount\_500]{%
    \includegraphics[width=0.48\linewidth]{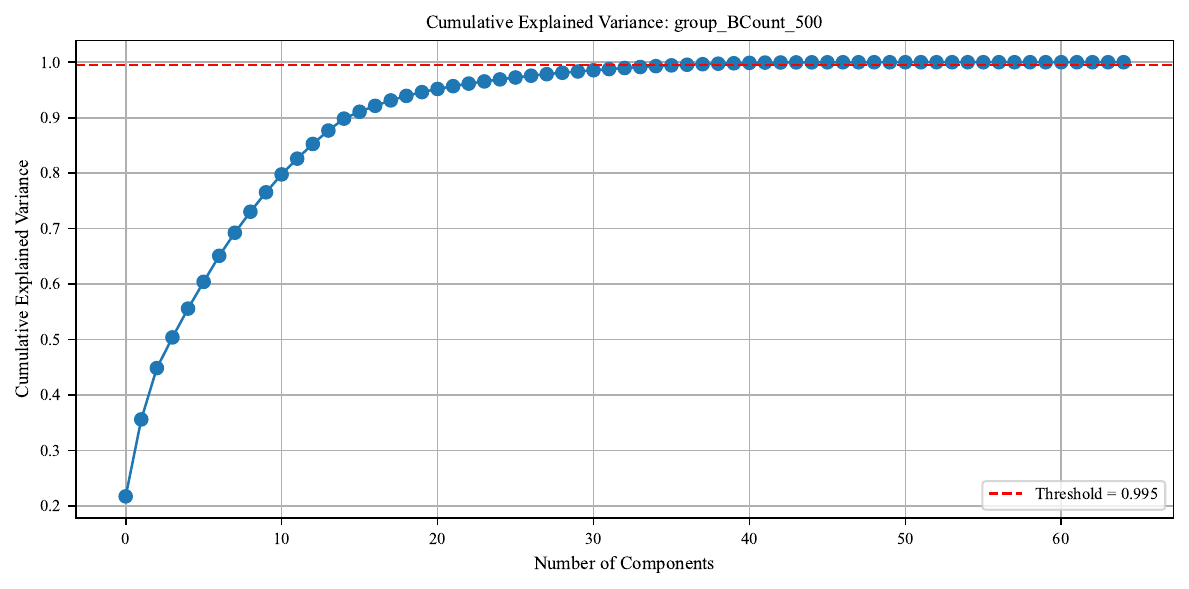}%
    \label{fig:3b}}
\caption{Cumulative explained variance for different feature groups.}
\label{fig:3}
\end{figure}

The proportion of explained variance was used to determine dimensionality, reflecting how much of the original dataset’s variance is retained \citep{raykov2014population}. To ensure high information retention, a 99.5\% explained variance threshold was adopted. Figure~\ref{fig:3} illustrates two examples: the \texttt{group\_accessibility} variables retained all 10 dimensions, while \texttt{group\_BCount\_500} was reduced from 65 to 37 dimensions (56.9\% retention).

The PCA process was uniformly applied across all 51 feature groups using the same variance threshold. Consequently, the number of features was reduced from 888 to 595, yielding a more compact and informative dataset for subsequent model development.

\subsection{Random Forest and Quantile Random Forest models}

Random Forest (RF) and Quantile Random Forest (QRF) models are supervised learning algorithms that relate a response variable $Y_i$, e.g. AADT for observation $i$, $i=(1,\dots,n)$, to a $k$ dimensional covariate vector $X_i=(X_{i1},\dots,X_{ik})$, which contains a set of variables (or features) that that we believe informative for prediction of $\hat{Y}_i$.

We use the RF algorithms to calculate our point predictions and predictions intervals. A RF is an ensemble based bagging, or bootstrap aggregation, method that averages across regression trees to form predictions (for a brief overview see \citep{Hastieetal:2009}). Averaging is applied to multiple trees fitted on random subsets of the data to reduce variance and avoid overfitting. 

Thus, given covariates $X$, and total number of trees $T$, $t=(1,\dots,T)$, each tree is trained on a random bootstrap sample on a random subset of features. The final RF point prediction $\hat{f}(x) $ is simply the average across all tree predictions given by
\begin{equation}
\hat{f}(x) = \frac{1}{T} \sum_{t=1}^{T} f_t(x)
\label{eq:average_function}
\end{equation}
where $f_t(x)$ is the prediction value from the $t$-th tree.

RF thus yields point estimates for each $i$. To predict a distribution of possible outcomes, we use the closely related QRF method to predict conditional quantiles (see \citep{Meinshausen2006}). For a given quantile \( q \in (0, 1) \), the model estimates:
\begin{equation}
\hat{Q}_q(x) = \inf \left\{ y \in \mathbb{R} : \hat{F}(y \mid x) \geq q \right\}
\label{eq:quantile_inf}
\end{equation}
where $\hat{Q}_q(x)$ is the estimated $q$-th quantile of $Y$ conditional on $X = x$, and $\hat{F}(y \mid x)$ is the empirical cumulative distribution function (CDF) for $Y \mid X = x$. The conditional empirical CDF is constructed from all observations $y_i$ in the leaf nodes across all trees that the input $x$ reaches. The predicted quantile is the smallest $y$ such that the conditional CDF is at least $q$, e.g.
\begin{equation}
\hat{Q}_q(x) = \min \left\{ y_i : \hat{F}(y_i \mid x) \geq q \right\}
\label{eq:quantile_min}
\end{equation}

The QRF method thus returns an interval that allows for uncertainty in prediction, and crucially for our application, provides estimates of lower and upper bounds for AADT. Furthermore, QRF does so without making strong distributional assumptions. Unlike parametric models that impose strict distributional assumptions, QRF accommodates complex nonlinearities and variations across heterogeneous data. Such flexibility is valuable in transportation research, where traffic dynamics are shaped by intertwined geographic, demographic, and infrastructure factors. This capability not only helps capture fluctuations and extremes in traffic flow, but also enhances the adaptability and scalability of the Random Forest algorithm in big data contexts \citep{Genuer2017}.

\subsection{Hyperparameter Tuning, Evaluation, and goodness of fit metrics}
The Quantile Random Forest (QRF) algorithm contains several key hyperparameters. The dataset was divided into training (80\%) and testing (20\%) subsets, and two complementary optimization strategies were applied: Random Search and Bayesian Optimization. To evaluate model performance and assess goodness-of-fit, a number of point and interval estimation metrics were used, including pseudo-$R^2$, the Risk Assessment Index (RAI) and the Winkler Score (WS). For details see the appendix.   

\subsection{Feature Importance Metrics}

To interpret model behavior and identify key predictors influencing AADT estimation, two complementary feature importance measures were employed: Mean Decrease Impurity (MDI) and Permutation Feature Importance (PFI).  

MDI, a built-in metric in tree-based models, quantifies each feature’s contribution to reducing prediction error across all splits in the ensemble. It provides an efficient and interpretable means of ranking features, although it may introduce bias toward variables with many distinct values \citep{breiman2001random, strobl2008conditional, louppe2013understanding}.  

In contrast, PFI is a model-agnostic approach that assesses the performance drop when individual feature values are randomly permuted \citep{molnar2023relating, debeer2020conditional}. This method offers a more general measure of feature relevance and can capture nonlinear relationships beyond those reflected in impurity-based measures.  

Together, these techniques provide complementary insights—MDI identifies features most frequently used for predictive splits, while PFI quantifies their actual contribution to predictive performance. Detailed derivations, equations, and implementation procedures for both metrics are provided in the appendix.

\section{Results}

Following data preparation, 51 feature groups were constructed based on spatial and thematic similarity (see Table~\ref{tab:feature_groups} in the Appendix).  
Applying PCA to each group effectively reduced dimensionality while retaining key predictive information, decreasing the total number of features from 888 to 595.  
Groups related to business counts (\texttt{BCount}) and employment variables retained the highest number of principal components, reflecting their detailed internal variability and strong relevance to AADT prediction (Figure~\ref{fig:4}).

Model training and tuning were performed using Quantile Random Forests (QRF) with hyperparameters optimized via Bayesian Optimization. The best-performing model—based on the exponential accessibility metric with $\alpha = 1.5$—achieved RMSE = 0.8821, MAE = 0.6734, pseudo-$R^2$ = 0.5916, and MAPE = 10.38.  The coefficient of variation (CV) of the 50\% quantile prediction error was 84.5\% (log scale), indicating substantial variability consistent with heterogeneous traffic patterns, while the CV of interval width (24.25\%) confirmed stable prediction uncertainty.

Feature importance was assessed using both Mean Decrease Impurity (MDI) and Permutation Importance (PFI). Both metrics consistently identified road-related features (\texttt{group\_road\_PC1} and \texttt{group\_road\_PC4}) as dominant predictors, alongside transport accessibility (\texttt{group\_transport\_1600\_PC1}, \texttt{group\_transport\_3200\_PC1}) and land-use characteristics (\texttt{group\_Class10\_code\_PC1}).  Although minor ranking differences occurred between MDI and PFI (Figures~\ref{fig:5}–\ref{fig:6}), eight of the ten top-ranked features overlapped, demonstrating consistency and robustness in feature relevance.

Error analysis (Figures.~\ref{fig:7}–\ref{fig:8}) showed that most predictions clustered closely around the true values and within the prediction intervals, indicating reliable model performance. Outliers, primarily in high-AADT ranges, exhibited larger deviations, reflecting data heterogeneity and potential effects of extreme values. Overall, the QRF model demonstrates strong generalization capability and interpretable feature behavior for interval-based AADT prediction.  
Full quantitative and graphical details are provided in the appendix.

\begin{figure*}[!t]
\centering
\captionsetup[subfigure]{labelformat=default}  
\subfloat[Geographic pattern of AADT prediction errors at the median (50th percentile)]{%
    \includegraphics[width=3in]{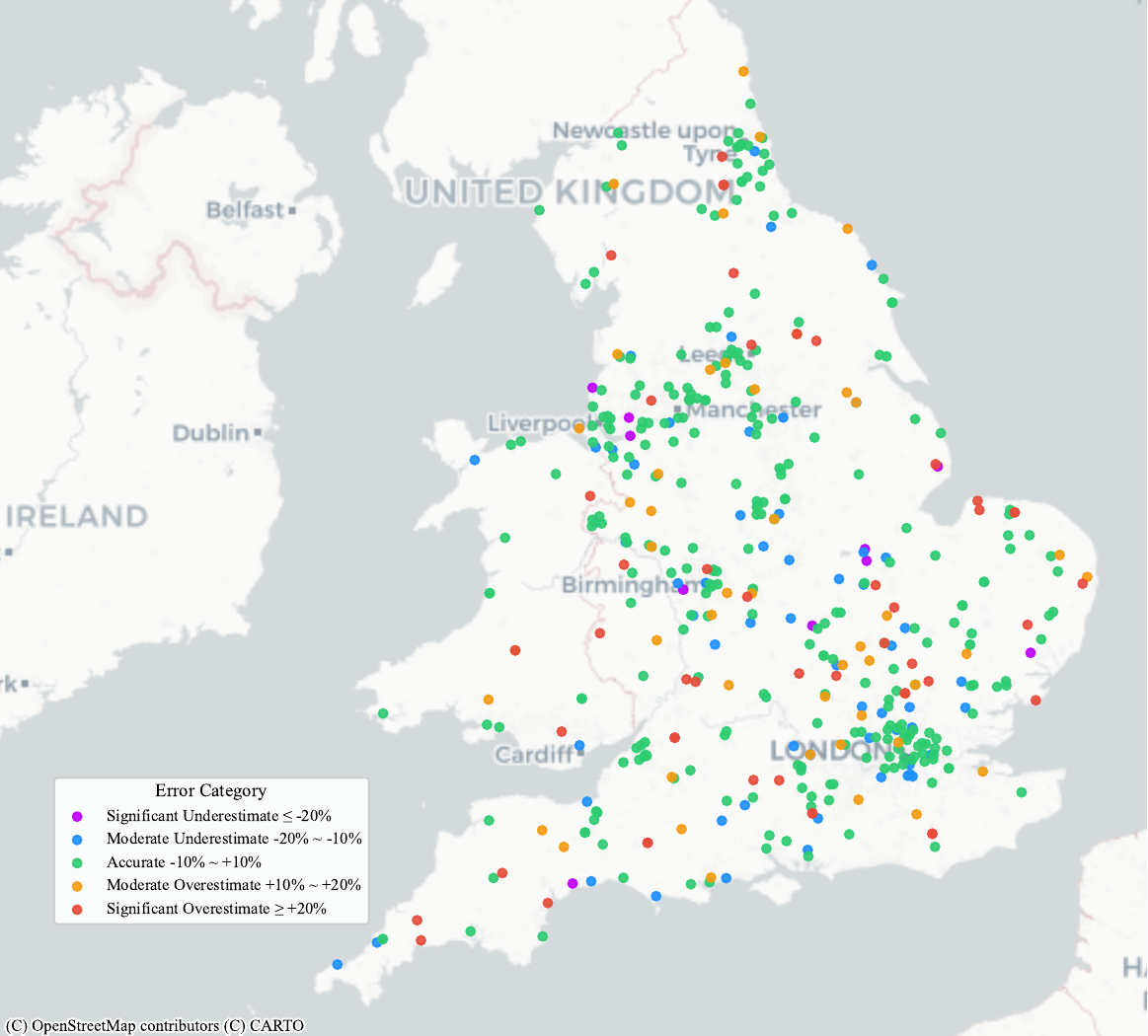}%
    \label{fig:9a}}
\hfill
\subfloat[Geographic variation in QRF prediction interval widths (log scale)]{%
    \includegraphics[width=3in]{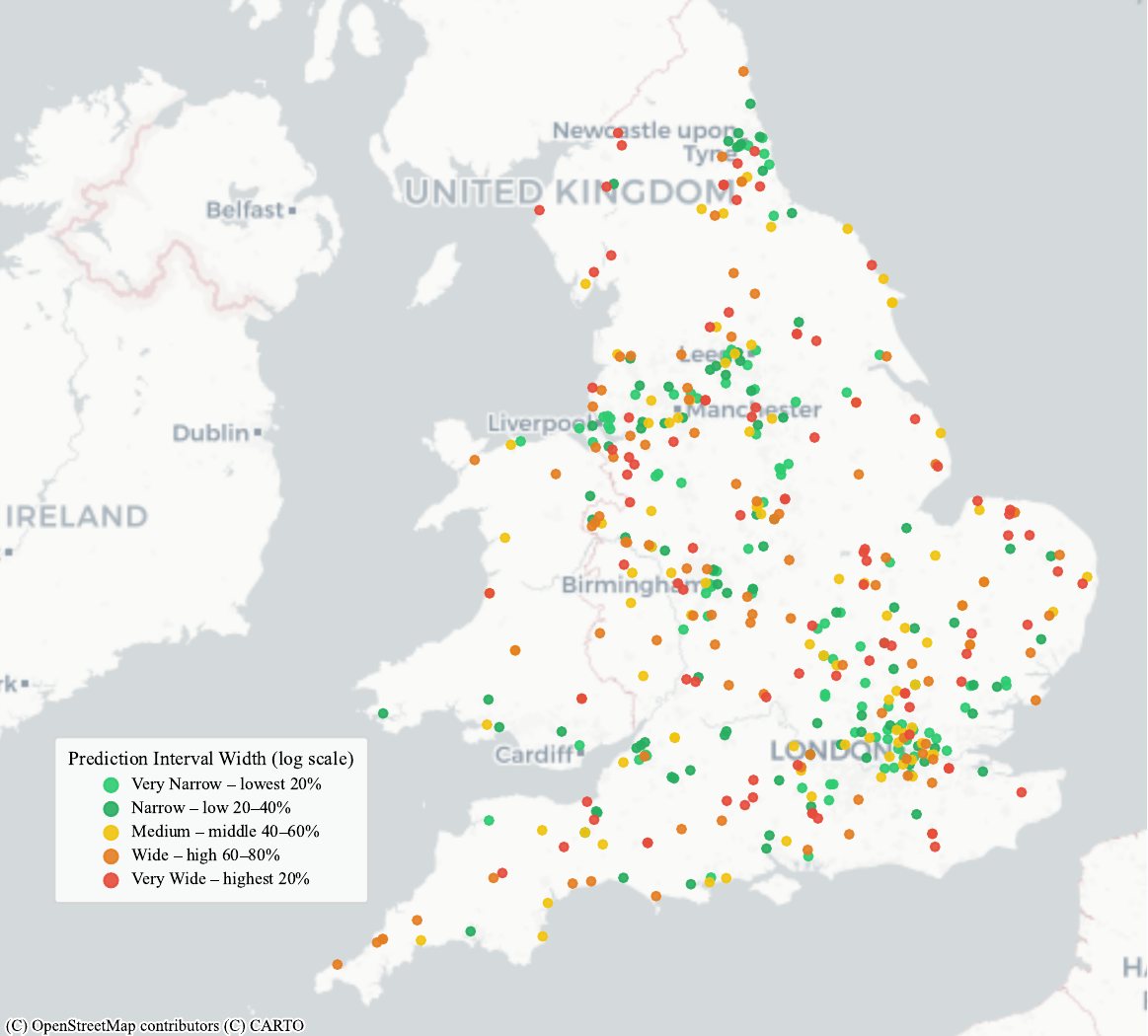}%
    \label{fig:9b}}
\caption{Spatial patterns of AADT prediction: 50\% quantile prediction error magnitudes and interval widths.}
\label{fig:9}
\end{figure*}

To enhance the interpretability of the model outputs, we visualise the 50\% quantile prediction results and the corresponding prediction intervals on geographic maps. Figure~\ref{fig:9a} illustrates the spatial distribution of the 50\% quantile prediction errors across England and Wales. Most of the prediction points in green had low margins of error, indicating high model accuracy. Moreover, the results demonstrate that prediction accuracy tends to improve in areas closer to large cities such as London and Liverpool. This suggests that the model performs better in regions with higher data density and more complex urban features. 

In addition, the prediction intervals are classified into five equal-sized groups based on their widths: very narrow, narrow, medium, wide, and very wide. This quantile-based grouping ensures an even distribution of prediction points across uncertainty levels. Figure~\ref{fig:9b} displays the spatial distribution of these interval width categories. The results are consistent with the 50\% quantile prediction error analysis. It also shows that locations of AADT observed point nearer to large cities tend to exhibit narrower prediction intervals, higher model confidence and reduced uncertainty in those areas.

Thus, the maps shown in Figure \ref{fig:9} reveal an interesting spatial pattern: links close to cities and populations centres tend to have narrower interval widths than links in rural and remote areas. This pattern can be verified by regressing interval width on the log of a zone level population accessibility index that we used as a predictor, given by
\begin{equation}\label{rg}    
\rho^G_i = \frac{1}{n} p_i \sum_{j=1}^n \frac{p_j}{d_{ij}^\alpha}.
\end{equation}
where $p$ denote zone population and $d$ is inter-zonal distance. The regression produces an $R^2$ value of 0.159 and a statistically beta coefficient of $-0.638$, indicating that interval width is increasing in less accessible places.

To explore the nature of this effect further we conducted a series of regressions on our interval estimate results. The two most pertinent findings are shown in Figure \ref{regscat} below.     

\begin{figure}[!t]
    \centering
    \captionsetup[subfigure]{labelformat=default}  
    \subfloat[AADT interval width versus AADT point estimate.]{%
        \includegraphics[width=0.48\linewidth]{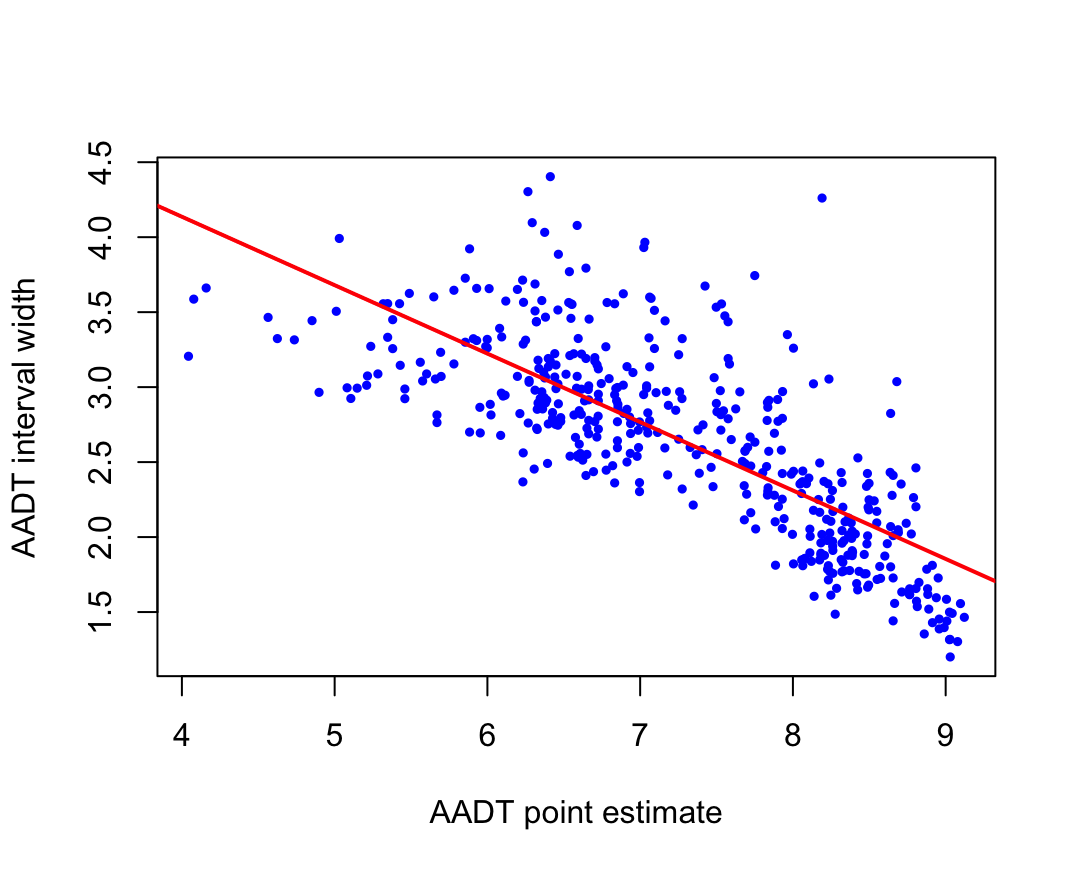}%
        \label{regscat:sub1}}
    \hfill
    \subfloat[AADT interval width versus AADT error.]{%
        \includegraphics[width=0.48\linewidth]{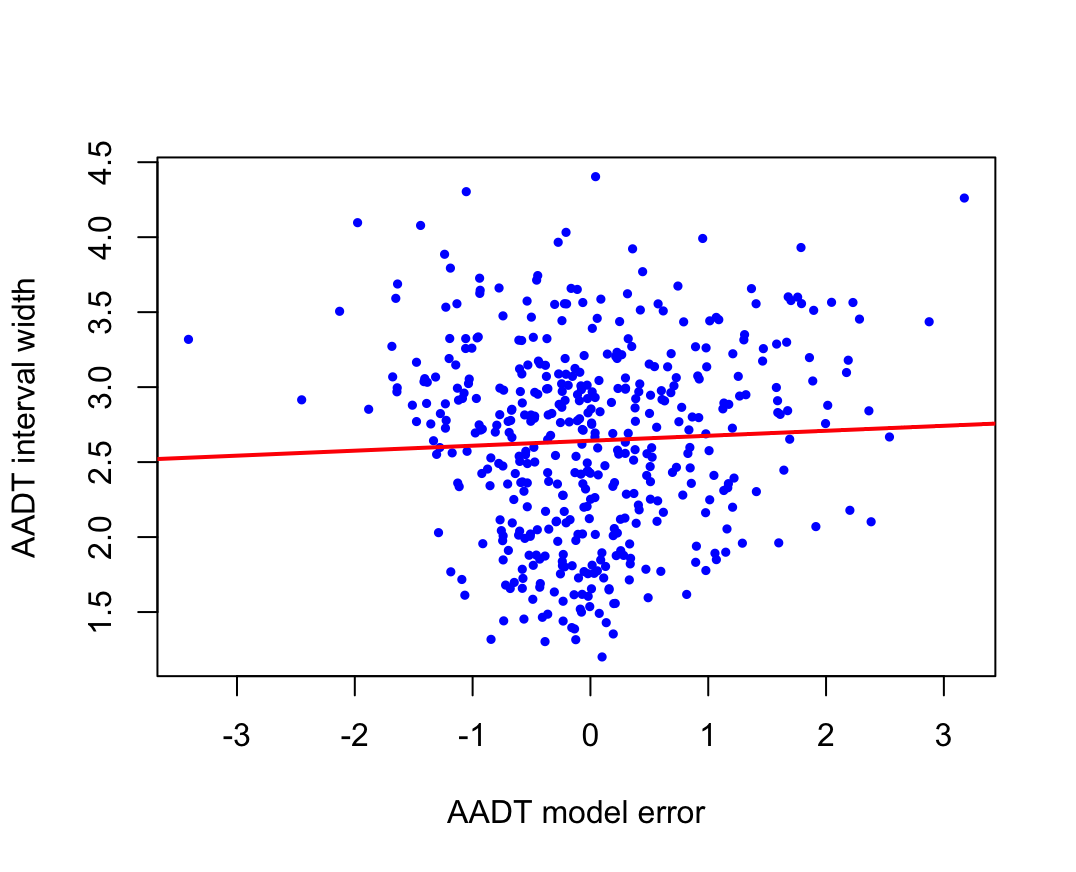}%
        \label{regscat:sub2}}
    \caption{Scatterplots with regression line fits: 
    (a) $R^2=0.591$, $\hat{\beta}_p=0.457$ (0.018), and 
    (b) $R^2=0.002$, $\hat{\beta}_e=0.033$ (0.033).}
    \label{regscat}
\end{figure}

Figure \ref{regscat:sub1} shows that interval width is strongly decreasing with the size of the point estimate. In other words, links with wider intervals tend to have lower level of traffic. Since heavily used links tend to be found in and near cities, this explains what lies behind the spatial patterns we observe in Figure \ref{regscat} above. Moreover, the relationship shown in Figure \ref{regscat:sub1} appeals to intuition, since more heavily used links are likely to have more stability and less variability due to recurrent patterns of use.

Interestingly, as shown in Figure \ref{regscat:sub2}, interval width is not associated with the value of the model error term. This result indicates that although interval width size is decreasing with the point estimate, the estimates themselves remain unbiased (e.g. there is no systematic change in the value of the error).
 
For AADT prediction, the implications are that less densely used links have inherently greater variability in traffic volumes and are therefore harder to predict, but that the predictions we have for these links are still generated with zero bias. This underscores the value of procuring interval rather than point predictions, in the sense that that it is particularly useful to have an interval for smaller capacity links that have greater variability in volumes.

Finally, this section reports the metrics used to evaluate model performance. Table~\ref{tab:model_performance} summarizes the values of the key evaluation indicators. The Prediction Interval Coverage Probability (PICP) reaches 88.22\%, slightly above the predefined threshold of 85\%, indicating that most true AADT values fall within the prediction intervals and reflecting the model’s moderate conservatism. The Normalized Average Width (NAW) is 0.23, suggesting that the intervals are of appropriate compactness—neither excessively wide nor overly narrow. The Risk Assessment Index (RAI) is 3.68, demonstrating a reasonable trade-off between predictive accuracy and interval coverage. 

\begin{table}[!t]
  \centering
  \setlength{\abovecaptionskip}{2pt} 
  \caption{Model performance.}
  \label{tab:model_performance}
  \renewcommand{\arraystretch}{1.2} 
  \begin{tabular}{l
                  S[table-format=2.2] 
                  S[table-format=1.2] 
                  S[table-format=1.2] 
                  S[table-format=4.2]}
    \hline
    {Metric} & {PICP (\%)} & {NAW} & {RAI} & {WS} \\
    \hline
    Result & 88.22 & 0.23 & 3.68 & 7468.47 \\
    \hline
  \end{tabular}
\end{table}

Thus, our AADT interval prediction model is good at minimizing the interval width and maintaining sufficient coverage. However, Winkler's score (WS) of 7,468.47 is relatively large. This is the inherent challenge of long-tailed distribution characteristics of traffic flow data. Considering that the observation range of traffic flow data is very large, the WS here is reasonable rather than indicative of a flaw in model validity. 

At this time, 60\% of the available features are randomly considered per tree (\texttt{max\_features = 0.6}) for each node split, which facilitates the identification of optimal split points while mitigating the risk of overfitting. The final model comprises 193 decision trees (\texttt{n\_estimators = 193}), with a maximum tree depth of 48 (\texttt{max\_depth = 48}). To further regulate model complexity, the minimum number of samples required to split an internal node is set to 16 (\texttt{min\_samples\_split = 16}), and the minimum number of samples required to form a leaf node is set to 8 (\texttt{min\_samples\_leaf = 8}). These hyperparameters were carefully tuned to achieve an optimal balance between model complexity and generalization performance.

Overall, the results demonstrate that the model is well suited for interval AADT prediction, achieving high predictive accuracy while explicitly accounting for uncertainty. Moreover, the approach provides a reliable framework for uncertainty quantification in highly variable traffic flow data, yielding statistically meaningful confidence bounds that can support transportation infrastructure planning and capacity evaluation.

\section{Applications in transportation engineering}

In this section we present indicative calculations to demonstrate the value of interval estimation of AADT for transportation engineering and planning. Using the interval estimates for the 450 test observations we calculate a discrete approximation to the total proportional change in AADT over the interval width using
\begin{equation}
\label{eq:dlog}
d \log y_i \;\approx\; 
\frac{\hat{y}_i^{\max} - \hat{y}_i^{\min}}
     {\hat{y}_i^{\text{med}}} .
\end{equation}

We then show what these differentials imply for forecasts of congestion and collision risk. 

\begin{figure}[!t]
    \centering
    \includegraphics[width=3.5in]{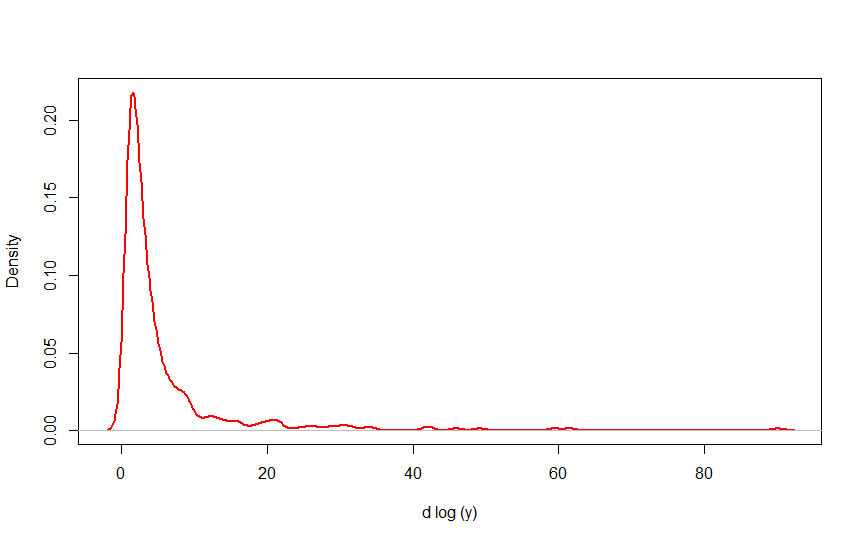}
    \caption{Proportional change in AADT over interval widths for the test data.}
    \label{dlogy}
\end{figure}

Figure \ref{dlogy} shows a density of the proportional changes in AADT over interval widths for the test data. The mean value is 5.509 and the standard deviation is 8.781. The figure shows that much of the mass is concentrated in the region $d \log y \leq 20$, but with considerable uncertainty for a small number of links (8) with proportional changes across the width in excess of 40.       

We remove the 8 extreme values showing extreme uncertainty from the right-hand-side of the distribution and make our calculations using the remaining 443 values. 

\subsection{Quantifying travel time variance and congestion costs}

We first use our QRF interval estimates to represent variance in travel times on a representative congestible link. We do so using the well known Bureau of Public Roads (BPR) congestion technology function given by
\begin{equation}
\label{eq:tformula}
t = t_f \left[ 1 + \alpha 
\left( \frac{q}{q_k} \right)^{\beta} \right] ,
\end{equation}

where $t$ is the travel time on the link, $t_f$ is the freeflow travel time, $q$ is traffic flow, $q_k$ is the technical capacity of the link, and $\alpha$ and $\beta$ are parameters that determine the congestion technology. The typical parameterisation of the function is $\alpha = 0.15$ and $\beta = 4$, and these are the values we use here (for a review of the BPR function see \citep{SV2007}). 

We make our calculations for a representative link that is 10 miles in length and in which the free flow speed is 40 mph. The free flow travel time $t_f$ is therefore 15 minutes, and we scale the flow $q$ by $d \log y$.

\begin{figure}[!t]
    \centering
    \includegraphics[width=\linewidth]{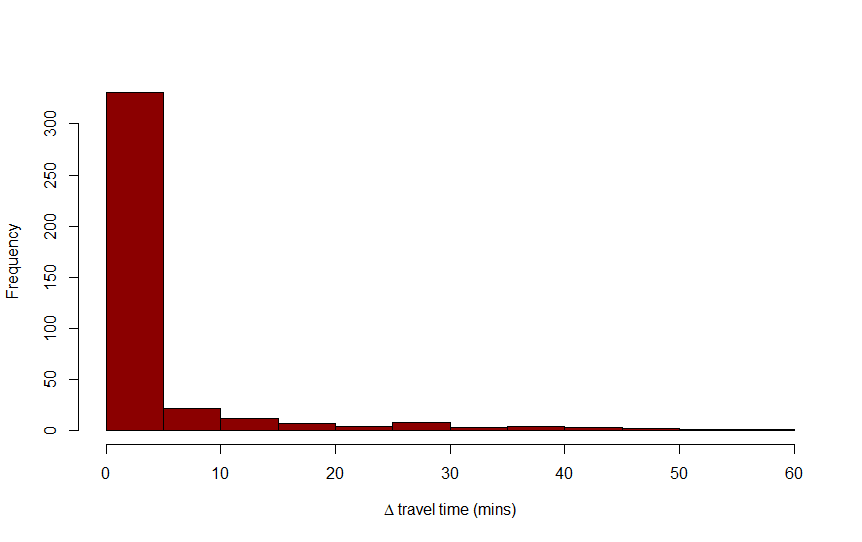}
    \caption{Absolute change in travel time over interval widths for the test data.}
    \label{dt}
\end{figure}

Figure \ref{dt} shows the difference in travel time from the BPR calculations between AADT volumes close to free-flow conditions, $q = 20$ and $q_k = 100$, and maximum AADT $q_{\max} = q (1 + d \log y)$ and $q_k = 100$. To aid visual interpretation, the right-hand side of the figure has been truncated at $\Delta t = 60 \text{ mins}$.

The figures shows that for many links, the difference in travel times between maximum and minimum estimated AADT appears relative small. For 275 of the 450 links the change in travel time is $\leq 1.5$ mins (or 10\% of the freeflow time). Note however, that we are making calculations just for one representation link in this case, and with additive excess travel time for a trip over multiple links, these small changes could still amount to substantial congestion costs.

For other links shown in the diagram the difference in travel time under maximum and minimum AADT is substantial. A difference of 50\% or more in travel times is forecast for 104 links and a doubling or more for 85 links. These results underscore the value of interval estimation in identifying links with characteristics that suggest substantial variance in travel times, potentially providing valuable insights that could help address bottlenecks in the network.      

\subsection{Assessing the risk of collisions}

The relationship between road traffic collisions and speeds has been studied extensively in the literature (see for example \citep{AARTS2006215}). Although complex and nuanced, \citet{elvik2004speed} find that a simple power law can provide a good approximation to the change in relative risk ($r$) induced by a change in speed ($v$), e.g.
\begin{equation}
\label{eq:delta_r}
\Delta r = \frac{C_A}{C_B} 
= \left( \frac{v_A}{v_B} \right)^{\beta} ,
\end{equation}

where $C_B$ is the number of collisions at the base (or ``before'') speed, $C_A$ is the number of collisions at the new (or ``after'' speed), and $\beta$ is a parameter governing the power law relationship.  

Using the speed values calculated via the BPR function above, with an assumed base speed of $v_B=40$ mph, along with $\beta$ parameter values from \citet{elvik2004speed}, we analyse variance in the risk of road traffic collision (RTCs) using our interval AADT estimates. The results are shown in table 3. 

\begin{table}[!t]
    \centering
    \setlength{\abovecaptionskip}{2pt} 
    \caption{Road traffic collision risk calculations.}
    \label{tab:my_label}
    \begin{tabular}{lcccc}
        \toprule
        \textit{Collision type} & $\beta$ & Mean $\Delta r$ & Var($\Delta r$) & IQR($\Delta r$) \\
        \midrule
        Fatal   & 3.6 & 0.665 & 0.140 & 0.614 \\
        Serious & 2.4 & 0.709 & 0.129 & 0.479 \\
        Slight  & 1.2 & 0.778 & 0.104 & 0.282 \\
        \bottomrule
    \end{tabular}
\end{table}

The table shows that for collision planning purposes the AADT interval estimates demonstrate considerable uncertainty in the relative risk predicted for links. The mean change in relative risk using a point estimate (e.g. the mean) is 0.665, 0.709 and 0.778 for fatal, serious and slight injury collision respectively. However, the interquartile range calculations underscore the high degree of variance that exists around these point estimates. Use of interval AADT estimates can therefore be used by traffic engineers and planner to identify both high risk and volatile points on the network.            

\section{Discussion}

\subsection{Summary of Key Findings}

Table~\ref{tab:model_performance} above presented the results of key performance indicators (RAI, PICP, NAW, and WS), showing values within a reasonable range and thus verifying the robustness and reliability of our interval prediction model. Notably, when applied to large-scale datasets, the model achieves a good balance between high coverage (88.22\%) and narrow prediction intervals (NAW of 0.23). Setting the same 0.5 weight for PICP and NAW helps achieve this balance when calculating RAI(3.68). Although the WS value is relatively high, given the high-dimensionality of the traffic flow data set, it is still within the acceptable range. These findings underscore the model’s ability to balance coverage and interval width, thereby ensuring both the accuracy and adaptability of the predictions.

A major contributor to these favorable outcomes is the effective dimensionality reduction achieved through PCA. By identifying and retaining the most informative principal components, PCA enhanced both the predictive accuracy and computational efficiency of the model. The model itself was developed using the QRF algorithm, which is particularly suited for generating interval predictions. By producing robust AADT estimates with explicit uncertainty bounds, this approach serves as a practical tool to support traffic management and infrastructure planning.

\subsection{Comparison with other relevant studies}

The model of \citet{aprillia2021statistical}, under a 5\% variance threshold, reached high PICP values of up to 95.2\% with impressively low NAW values (as small as 0.095). Our model demonstrates competitive performance by achieving a PICP of 88.22\% at a fixed confidence level of 85\%, with a NAW of 0.23. While our coverage is slightly lower, our Winkler Score is substantially better at 7468.47, compared to the lowest value of \citet{aprillia2021statistical} which was above 277,000. Thus, compared with the work of \citet{aprillia2021statistical}, our model achieves a strong balance between interval sharpness and predictive accuracy.  It is also worth noting that traffic flow data tend to be more variable and less predictable than electric load data, which adds to the challenge.

\citet{vanStrien2024} developed a global AADT prediction model based on QRF which achieves a pseudo-$R^2$ of 0.74 and a prediction interval coverage of 92.1\% under a 90\% confidence level. Our study adopts a similar QRF model which is applicable to the high uncertainty and non-normal distribution characteristics of traffic flow data. Although the confidence level set in our study is relatively low (85\%), the actual coverage percentage still reaches 88.22\%. This indicates that the interval prediction of the model is still of practical significance. It is similar to the coverage percentage of 92.1\% of \citet{vanStrien2024} (The confidence level set is 90\%). While \citet{vanStrien2024} mainly focused on the coverage of confidence intervals, our analysis further introduces the NAW, RAI and WS indicators, which offers an important supplement for model evaluation. 

\section{Conclusions}

This study employs a QRF framework to produce interval predictions of AADT for minor roads in England and Wales. The proposed methodology is structured around five main stages:
\begin{enumerate}
    \item Data pre-processing to normalize, refine, filter and transform the data
    \item Principal Component Analysis (PCA) to achieve dimensionality reduction within high-dimensional feature groupings
    \item Hyperparameter tuning on training and test datasets to optimize the performance of the model
    \item Formulation of the QRF model for interval prediction
    \item Model evaluation was conducted using diagnostic metrics, including RAI, WS, PICP, and NAW.
\end{enumerate}

It is the first time that QRF has been applied to AADT interval prediction for minor roads in England and Wales, taking high-dimensional factors into account. In the paper we have demonstrated that the characteristics of this method with high prediction accuracy and robustness. Furthermore, we have shown the value that AADT interval estimation holds for transportation engineering and planning through some simple case study calculations.

\bibliographystyle{chicago}
\bibliography{AADT}
\pagebreak

\appendix
\section*{APPENDIX}
\section{Summary of Interval Prediction Studies}\label{LRS}
{\footnotesize
\renewcommand{\arraystretch}{1.2}
\setlength{\tabcolsep}{3pt}
\footnotesize
\renewcommand{\arraystretch}{1.05}
\setlength{\tabcolsep}{2pt}

\begin{longtable}{|p{0.16\columnwidth}|
                  p{0.23\columnwidth}|
                  p{0.20\columnwidth}|
                  p{0.33\columnwidth}|}
\caption{Summary of Interval Prediction Methods\label{tab:method_summary}}\\
\hline
\textbf{Author} & \textbf{Data} & \textbf{Model} & \textbf{Results} \\
\hline
\endfirsthead

\hline
\textbf{Author} & \textbf{Data} & \textbf{Model} & \textbf{Results} \\
\hline
\endhead

\hline
\multicolumn{4}{|c|}{\textbf{Statistical Prediction Methods}}\\
\hline
\cite{Botella2024} & MA of 15 studies (standardized mean differences) & FEM/REM with Hartung--Knapp adjustment & Compared CI and PI; demonstrated PI's role in reflecting heterogeneity.\\
\hline
\cite{Qi2022} & Strain of maize seeds & SN-PQ-Intervals & Unified pivotal method for interval estimation; validated via simulations.\\
\hline
\cite{IntHout2016} & Cochrane Database (2009--2013) & REMA (HKSJ, EB, PI) & 72.4\% of significant MA showed PIs crossing null; 20.3\% included opposite effects.\\
\hline
\cite{Xie2014} & Shanghai Composite daily prices (K-line) & GW-D3:D6-IPA & Improved interval prediction reliability (80\% for 2--3 trading days).\\
\hline
\cite{Yu2009} & Normal, Exponential, Weibull, Lognormal & Box-Cox Exponential Transformation & Precise PI; validated by simulation.\\
\hline

\multicolumn{4}{|c|}{\textbf{Machine Learning Methods}}\\
\hline
\cite{Dang2022} & Singapore electricity market & QRRF with VGGNet, ResNet, Inception, LSTM & Accurate \& reliable probabilistic forecasting.\\
\hline
\cite{Genuer2017} & Simulated (15M), Airline (120M) & Parallel RF, Online RF, Subsampling & Big Data RF: Accurate \& noise-robust.\\
\hline
\cite{Meinshausen2006} & Boston Housing, Ozone, Abalone, BigMac, Fuel & QRF & Conditional quantile estimation; strong prediction; noise-robust.\\
\hline

\multicolumn{4}{|c|}{\textbf{Hybrid Statistical--Machine Learning Prediction Methods}}\\
\hline
\cite{lai2022exploring} & UCI regression datasets (e.g., Boston Housing, Wine Quality) & Deep Neural Network -- Hybrid Loss Ensemble for Prediction Intervals & 95\% PICP with tighter MPIW; handled both uncertainty types.\\
\hline
\cite{niu2022combined} & EU ETS carbon trading prices & TVF-EMD + IS-MODA + Lognormal & High-accuracy PI for carbon prices (MAPE: 0.80\%--1.12\%).\\
\hline
\cite{aprillia2021statistical} & ISO-NE data & QRRF with WOA-DWT and RAI & Accurate probabilistic forecasting with risk assessment.\\
\hline
\cite{jiang2017semi} & Shanghai Stock Exchange Index & GARCH-QRRF with t/GED distributions & Superior to parametric methods; optimal with t/GED.\\
\hline
\cite{Splawinska2018} & Traffic volume: 86 stations (2000--2015) & ANN, Discriminant Analysis & Key factors for 100\% accurate seasonal classification.\\
\hline

\end{longtable}
\scriptsize
\noindent \textit{Notes:} REMA = Random-effects meta-analysis; FEM/REM = Fixed/Random Effects Models; 
CI/PI = Confidence/Prediction Interval; SN-PQ = Skew Normal Pivotal Quantity; 
GW-IPA = Grey Wrapping Interval Prediction Axioms; QRF = Quantile Regression Forests; 
RF = Random Forest; LSTM = Long Short-Term Memory; ANN = Artificial Neural Networks; 
MPIW = Mean Prediction Interval Width; GARCH = Generalized Autoregressive Conditional Heteroskedasticity; 
GED = Generalized Error Distribution; WOA = Whale Optimization Algorithm; 
DWT = Discrete Wavelet Transform; RAI = Risk Assessment Index.
\normalsize
\renewcommand{\arraystretch}{0.95}
\setlength{\tabcolsep}{2.5pt}
\normalsize

\normalsize}

\section{Tuning Hyperparameters}
The Quantile Random Forest (QRF) algorithm contains several key hyperparameters: the number of trees (\texttt{n\_estimators}), maximum tree depth (\texttt{max\_depth}), minimum samples required for node splitting (\texttt{min\_samples\_split}), minimum samples per leaf node (\texttt{min\_samples\_leaf}), and the maximum number of features considered at each split (\texttt{max\_features}). Increasing \texttt{n\_estimators} typically improves model stability, while limiting \texttt{max\_depth} mitigates overfitting. The parameters \texttt{min\_samples\_split} and \texttt{min\_samples\_leaf} control tree complexity, and \texttt{max\_features} regulates feature selection at each split.  
Proper tuning of these parameters is essential to balance model accuracy and generalization \citep{probst2019hyperparameters}.

The dataset was divided into training (80\%) and testing (20\%) subsets, and two complementary optimization strategies were applied: Random Search and Bayesian Optimization. Both methods used five-fold cross-validation (\texttt{cv=5}), 150 iterations (\texttt{n\_iter=150}), and full CPU parallelization (\texttt{n\_jobs=-1}). Reproducibility was ensured with \texttt{random\_state=42}. The search space for key hyperparameters was defined as follows: \texttt{n\_estimators} $\in$ [30, 300], \texttt{max\_features} $\in$ \{\texttt{'sqrt'}, \texttt{'log2'}, 0.4, 0.5, 0.6\}, \texttt{max\_depth} $\in$ [10, 50], \texttt{min\_samples\_split} $\in$ [2, 20], and \texttt{min\_samples\_leaf} $\in$ [1, 15].

\subsection{Random Search}
Random Search samples parameter combinations from predefined ranges and evaluates model performance for each configuration \citep{nalatissifa2021customer}. Despite its stochastic nature, it often identifies near-optimal solutions efficiently and is well suited for large datasets \citep{castellanos2023improving, aprillia2021statistical, lima2021toward}. The \texttt{RandomizedSearchCV} function from \texttt{Scikit-Learn} was employed to tune the \texttt{Random\-Forest\-Quantile\-Regressor}, leveraging its simplicity, parallel computation, and robustness to local optima.

\subsection{Bayesian Optimization}
Bayesian Optimization refines hyperparameters by constructing a probabilistic surrogate model of the objective function and iteratively updating it to locate the most promising configurations \citep{joy2020fast}. Using Gaussian Process modeling, it efficiently explores the parameter space in high-dimensional or computationally expensive settings \citep{castellanos2023improving, aprillia2021statistical, masum2021bayesian}. The \texttt{BayesSearchCV} implementation from \texttt{scikit-optimize (skopt)} was applied to the same search space, employing identical iteration and cross-validation settings. The final QRF model was refitted using the optimal parameters identified through Bayesian optimization.

Together, these tuning procedures enabled efficient exploration of the hyperparameter space, reducing overfitting risk and improving predictive accuracy. Both methods are supported natively in \texttt{Scikit-Learn} and its extensions, providing a flexible and reproducible framework for model optimization \citep{castellanos2023improving}.

\section{Evaluation and Goodness of Fit Metrics}
\subsection{Point Prediction Evaluation Metrics}
The $R^2$ value defined in Eq.\eqref{eq:cpseudo-R^2-1} is standard in evaluating non-parametric models such as Random Forests or Quantile Regression Forests.  While it resembles the classical $R^2$ from linear regression, it does not carry the same interpretative meaning in terms of explained variance due to the non-linear and ensemble nature of the model. The coefficient of variation based on the point prediction model is calculated using Eq.\eqref{eq:cv_error}. This metric provides a standardized discrete metric and can also consistently evaluate the predictive uncertainty of different datasets or model configurations. We will use these metrics to evaluate the performance of 50\% quantile prediction result.

\begin{equation}
\label{eq:pseudo-R^2}
\hat{y}_i = \hat{q}_{0.5,i}
\end{equation}
\begin{equation}
\label{eq:cpseudo-R^2-1}
\text{Pseudo-}R^2 = 1 - \frac{\sum(y_i - \hat{q}_{0.5,i})^2}{\sum(y_i - \bar{y})^2}
\end{equation}
\begin{equation}\label{eq:cv_error}
    CV_{error} = \frac{\sigma_{error}}{\mu_{error}}\times 100\%
\end{equation}

\noindent
where $\sigma_{error}$ is the standard deviation of absolute prediction errors and $\mu_{error}$ is the mean absolute error.

\subsection{Interval Prediction Evaluation Metrics}
The Risk Assessment Index (RAI) and the Winkler Score (WS) are two widely used metrics for evaluating model performance in interval prediction \citep{aprillia2021statistical}. RAI integrates the Normalized Average Width (NAW) of the prediction intervals with the Prediction Interval Coverage Probability (PICP) to measure model reliability, balancing interval width and coverage using weights $w_1 + w_2 = 1$. WS assesses the accuracy of prediction intervals by penalizing intervals that fail to capture the true values while rewarding narrower intervals that do \citep{aprillia2021statistical}. In this study, the computation of RAI is presented in Eqs.~\eqref{eq:1}--\eqref{eq:4}, while the calculation of WS is given in Eq.~\eqref{eq:5}. A higher RAI value together with a lower WS score indicates superior predictive performance. To further test the superiority of the interval prediction method over the point prediction method, we also calculated the coefficient of variation of the interval width, as shown in Eq.~\eqref{eq:cv_width}. This metric quantifies the relative dispersion of the interval width. A smaller $CV_{\text{width}}$ implies a more uniform and stable interval prediction, which indicates a robust prediction model.

\begin{equation}
\label{eq:1}
NAW = \frac{1}{N} \sum_{i=1}^{N} 
\frac{U_i - L_i}{\operatorname{Range}(y)}
\end{equation}

\begin{equation}
\label{eq:2}
PICP = \frac{1}{N} \sum_{i=1}^{N} 
\mathbf{1}\!\left(y_i \in [L_i, U_i]\right) \times 100\%
\end{equation}

\begin{equation}
\label{eq:3}
RAI = w_1 \cdot NAW^{-1} + w_2 \cdot PICP
\end{equation}

\begin{equation}
\label{eq:4}
w_1 + w_2 = 1
\end{equation}

\begin{equation}
\label{eq:5}
WS =
\begin{cases}
(U_i - L_i) + \dfrac{2}{\alpha}(L_i - y_i), & y_i < L_i, \\[1.2ex]
(U_i - L_i), & L_i \leq y_i \leq U_i, \\[1.2ex]
(U_i - L_i) + \dfrac{2}{\alpha}(y_i - U_i), & y_i > U_i .
\end{cases}
\end{equation}

\noindent
where $L_i$ and $U_i$ are the lower and upper bounds of the prediction interval, respectively, and $\alpha = 1 - \text{coverage percent}$.

\begin{equation}
\label{eq:cv_width}
CV_{\text{width}} = 
\frac{\sigma_{\text{width}}}{\mu_{\text{width}}} \times 100\%
\end{equation}

\noindent
where $\sigma_{\text{width}}$ is the standard deviation of prediction interval widths and $\mu_{\text{width}}$ is the mean interval width.

\medskip
In this study, RAI is adopted as the primary criterion for selecting optimal hyperparameters during model tuning. The comparison between values obtained from Random Search and Bayesian Optimization determines which configuration is preferred, with the higher RAI indicating superior performance. In particular, under the predefined coverage of 85\%, the hyperparameter setting that yields the highest RAI is chosen. Model evaluation is then conducted based on this optimal configuration, and the detailed outcomes are reported in Sections~4 and 5. This approach offers a systematic framework for refining hyperparameters in predictive modeling.

\section{Feature Importance Metrics: Detailed Formulation}
\label{appendix:featureimportance}

\subsection{Mean Decrease Impurity (MDI)}

Mean Decrease Impurity (MDI) measures the contribution of each feature to the reduction in node impurity (e.g., Gini, entropy, or MSE) across all trees in an ensemble \citep{gwetu2019exploring, strobl2008conditional}.  
Following \citet{ishwaran2007variable, breiman2001random}, the average impurity reduction achieved by splits on feature $x_j$ defines its importance score:  

\begin{equation}
\label{eq:MDI}
\text{MDI}_j = 
\sum_{\text{splits on } x_j} 
\frac{N_t}{N} \cdot 
\bigl(\text{impurity}_{\text{parent}} 
- \text{impurity}_{\text{left}} 
- \text{impurity}_{\text{right}}\bigr)
\end{equation}

\begin{equation}
\label{eq:MDI2}
\text{impurity} = \text{MSE} 
= \frac{1}{N} \sum_{i=1}^{N} (y_i - \hat{y})^{2}
\end{equation}

\noindent
where $N_t$ is the number of samples at a given node, $N$ the total sample size, $y_i$ the observed values, and $\hat{y}$ the predicted values.  
MDI has been widely used across domains such as medicine, social sciences, and environmental modeling for identifying influential variables and improving model interpretability \citep{Liaw&Wiener2002, Pedregosa2011}.  
However, it can exhibit bias toward continuous or multi-level categorical features \citep{strobl2007bias}.

\subsection{Permutation Feature Importance (PFI)}

Permutation Feature Importance (PFI) is a model-agnostic approach that quantifies the importance of each variable by measuring the decline in predictive performance after random permutation of that feature \citep{debeer2020conditional, molnar2023relating}.  
For a trained model $f$, PFI for feature $X_j$ is computed as:

\begin{equation}
\label{eq:PFI}
\text{PFI}(X_j) = S(f, X, y) - S(f, X^{\pi_j}, y)
\end{equation}

\noindent
where $S(f, X, y)$ is the model's original performance (e.g., accuracy, F1-score, or RMSE), and $X^{\pi_j}$ is the dataset with the $j$-th feature permuted across samples.  
A larger performance drop indicates a more influential feature \citep{ramosaj2023consistent, afanador2013bootstrap}.  
Variants such as margin-based and AUC-based permutation importance enhance stability and robustness for imbalanced datasets \citep{yang2017margin, janitza2012auc}.  
PFI has been applied successfully across fields including bioinformatics and cybersecurity to identify features critical to prediction accuracy \citep{altmann2010corrected, abdelaziz2025enhancing}.

\section{Detailed PCA, Model, and Feature Importance Results}
\label{appendix:results}

\subsection{PCA and Dimensionality Reduction}
This section reports the full PCA dimensionality reduction results referenced in the main paper. 
Table~\ref{tab:pca_results} presents the original number of dimensions, the retained dimensions, 
and the retention rates for each feature group.

\begin{table}[H]
\centering
\begin{table}[H]
\caption{PCA Dimensionality Reduction Results for rhoEa15}
\label{tab:pca_results}
\centering
\footnotesize
\renewcommand{\arraystretch}{1.1}
\begin{tabular}{lccc}
\toprule
\textbf{Group Name} & \textbf{Original Dim.} & \textbf{Remaining Dim.} & \textbf{Retention Rate} \\
\midrule
\multicolumn{4}{l}{\textbf{Basic Geographic and Accessibility Features}} \\
group\_latitude      & 1  & 1  & 100.0\% \\
group\_longitude     & 1  & 1  & 100.0\% \\
group\_Class10\_code & 1  & 1  & 100.0\% \\
group\_accessibility & 10 & 10 & 100.0\% \\
group\_road          & 6  & 4  & 66.7\% \\
group\_ports         & 6  & 5  & 83.3\% \\
group\_airports      & 6  & 4  & 66.7\% \\
group\_lag           & 1  & 1  & 100.0\% \\
\midrule
\multicolumn{4}{l}{\textbf{Buffer Zone Features - 500m}} \\
group\_BCount\_500   & 65 & 37 & 56.9\% \\
group\_junc\_500     & 1  & 1  & 100.0\% \\
group\_transport\_500& 4  & 4  & 100.0\% \\
group\_employment\_500& 51 & 36 & 70.6\% \\
group\_population\_500& 10 & 7 & 70.0\% \\
group\_vehicles\_500 & 7  & 5  & 71.4\% \\
group\_earnings\_500 & 4  & 4  & 100.0\% \\
\midrule
\multicolumn{4}{l}{\textbf{Summary}} \\
Total Feature Groups        & \multicolumn{3}{c}{51} \\
Total Original Dimensions   & \multicolumn{3}{c}{888} \\
Total Remaining Dimensions  & \multicolumn{3}{c}{595} \\
Dimension Retention Rate    & \multicolumn{3}{c}{67.0\%} \\
\bottomrule
\end{tabular}
\end{table}

\end{table}

Features were grouped into 51 categories based on type and spatial proximity.
PCA was then applied within each group, reducing the original 888 features to 595 principal components while retaining 99.5\% of the total explained variance. Groups related to \texttt{BCount} and employment variables exhibited the highest retained dimensionality, indicating substantial inherent variability across these domains. Detailed descriptions of feature groupings and PCA outcomes are provided in the supplementary material.
Figure~\ref{fig:4} highlights the ten groups with the most retained components.

\begin{figure}[!t]
    \centering 
    \includegraphics[width=5.5in]{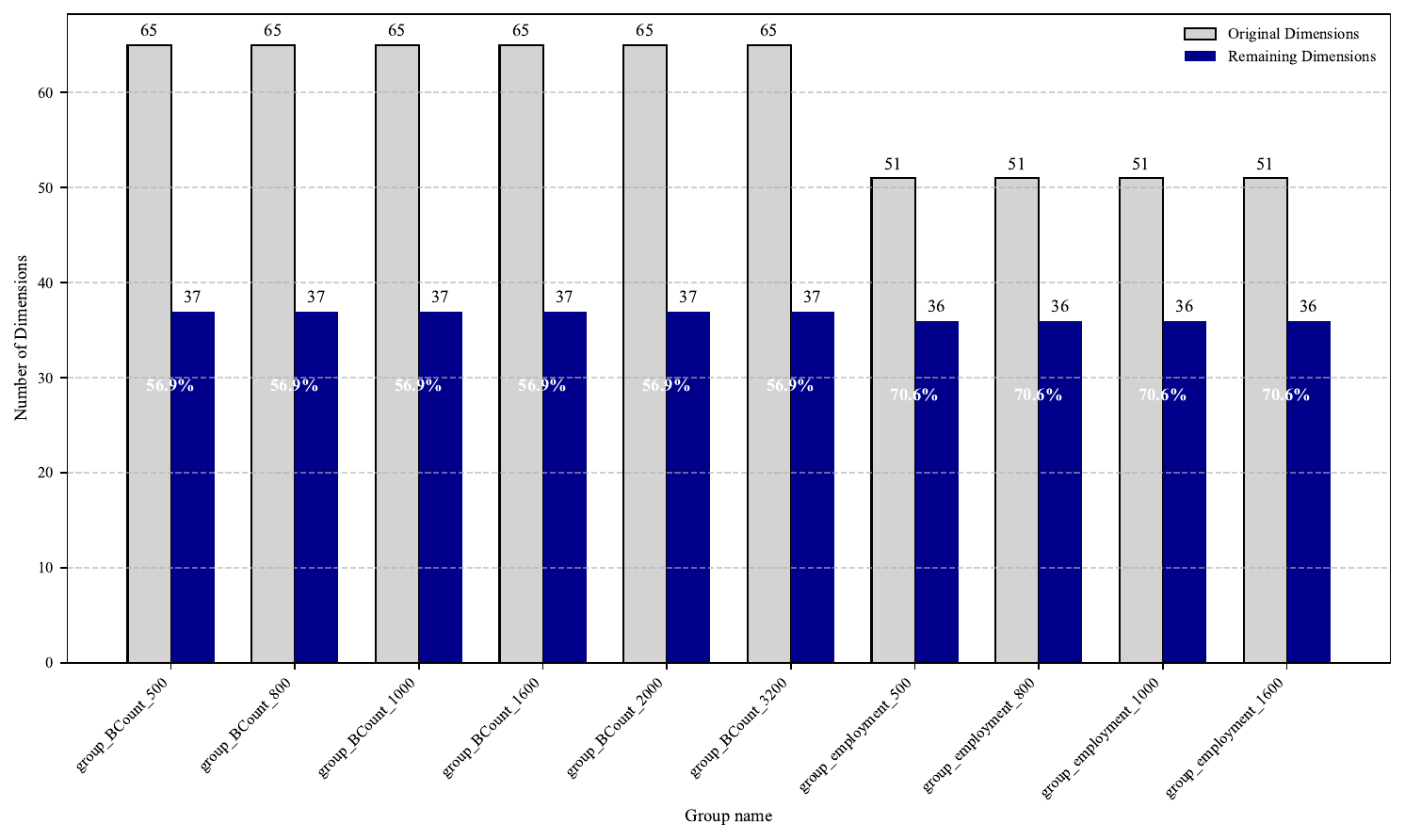}
    \caption{Top 10 groups by dimension after PCA.}
    \label{fig:4}
\end{figure}

\subsection{Model Optimization and Performance}

The QRF model was trained using 80\% of the data and validated on 20\%.  
Bayesian Optimization yielded the best configuration with the exponential accessibility metric at $\alpha = 1.5$.  
The optimal model achieved RMSE = 0.8821, MAE = 0.6734, pseudo-$R^2$ = 0.5916, and MAPE = 10.3788.  
Optimal hyperparameters included bootstrap enabled, \texttt{max\_depth} = 15, \texttt{max\_features} = 0.5, \texttt{min\_samples\_leaf} = 5, \texttt{min\_samples\_split} = 5, and 391 estimators.

The coefficient of variation of the 50\% quantile point prediction error ($CV_{error}$, Eq.~\ref{eq:cv_error}) was 84.50\% (log scale), while the coefficient of variation of interval width ($CV_{width}$, Eq.~\ref{eq:cv_width}) was 24.25\%.  
These values indicate heterogeneous yet stable predictive performance, justifying the selected features for interval prediction.

\subsection{Feature Importance and Interpretation}

\begin{figure}[H]
    \centering 
    \includegraphics[width=5.5in]{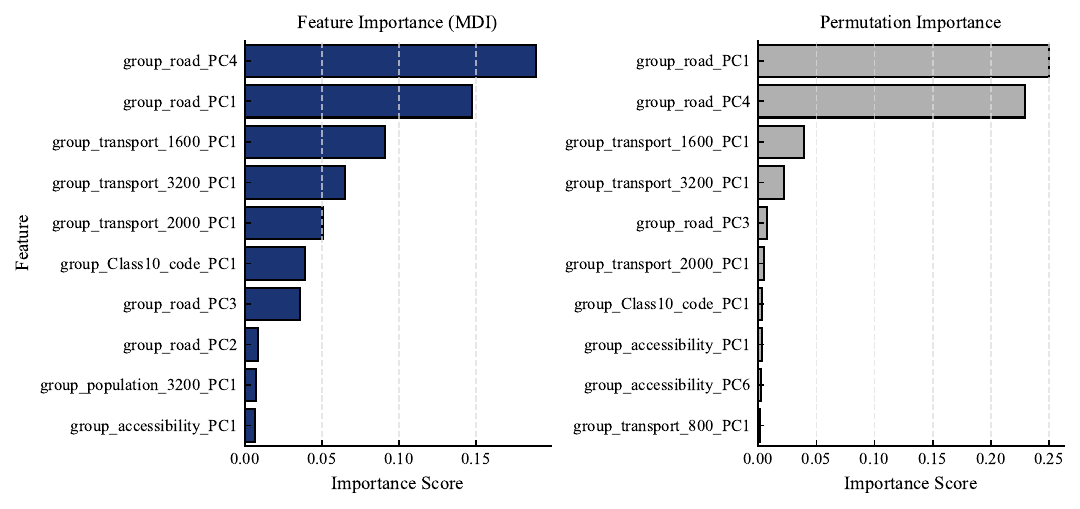}
    \caption{Top 10 features for AADT prediction using QRF with MDI and permutation importance.}
    \label{fig:5}
\end{figure}

\begin{figure}[H]
    \centering 
    \includegraphics[width=5.5in]{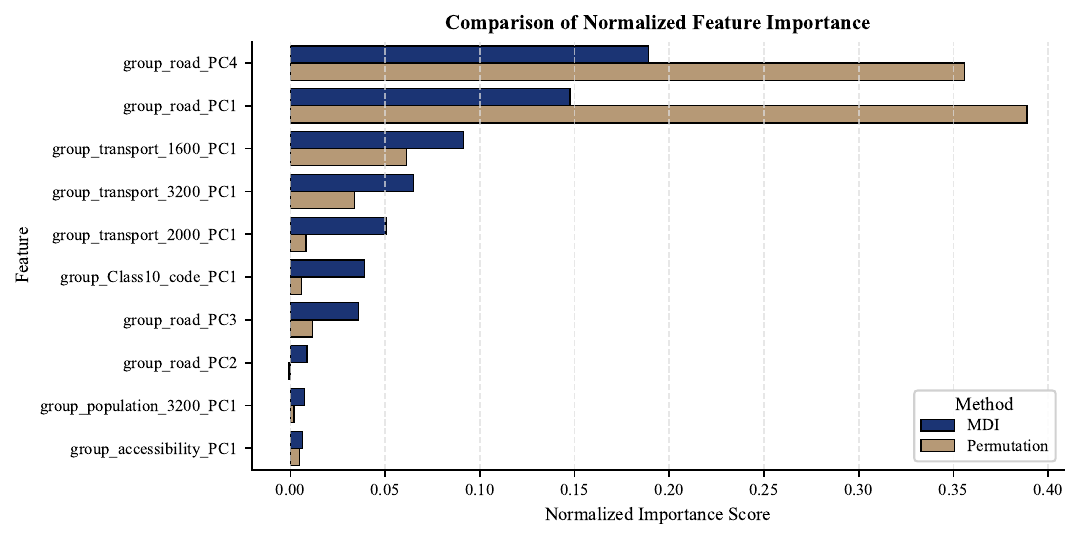}
    \caption{Comparison of normalized feature importance scores from MDI and permutation importance.}
    \label{fig:6}
\end{figure}

Feature importance derived from both MDI and PFI consistently highlighted road-related (\texttt{group\_road\_PC1}, \texttt{group\_road\_PC4}) and transport accessibility variables (\texttt{group\_transport\_1600\_PC1}, \texttt{group\_transport\_3200\_PC1}, \texttt{group\_transport\_2000\_PC1}) as key predictors.  
Land use (\texttt{group\_Class10\_code\_PC1}), accessibility (\texttt{group\_accessibility\_PC1}), and population density (\texttt{group\_population\_3200\_PC1}) also ranked among the top contributors.  
Figures~\ref{fig:5} and \ref{fig:6} compare importance scores across both methods, showing eight overlapping features in the top-10 list, underscoring their robustness.

\subsection{Error Analysis}

\begin{figure}[H]
    \centering 
    \includegraphics[width=5.5in]{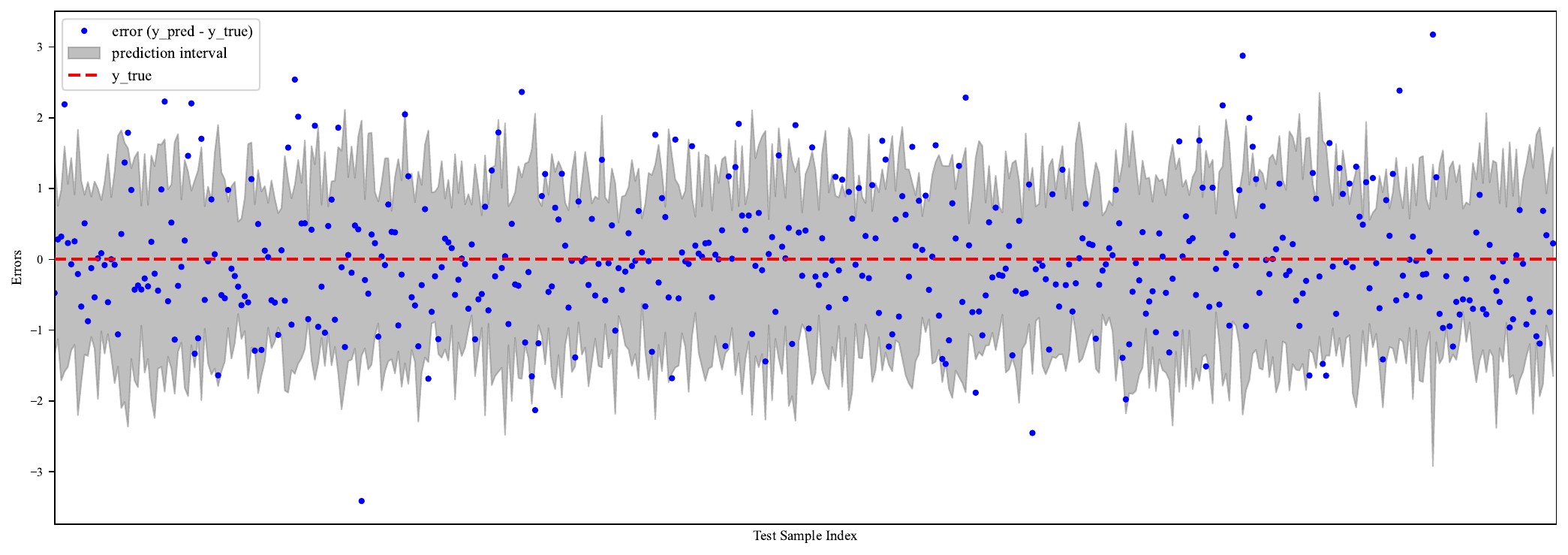}
    \caption{Prediction errors versus true values. Blue points denote individual prediction errors, with grey intervals showing uncertainty bounds.}
    \label{fig:7}
\end{figure}

\begin{figure}[H]
    \centering 
    \includegraphics[width=5.5in]{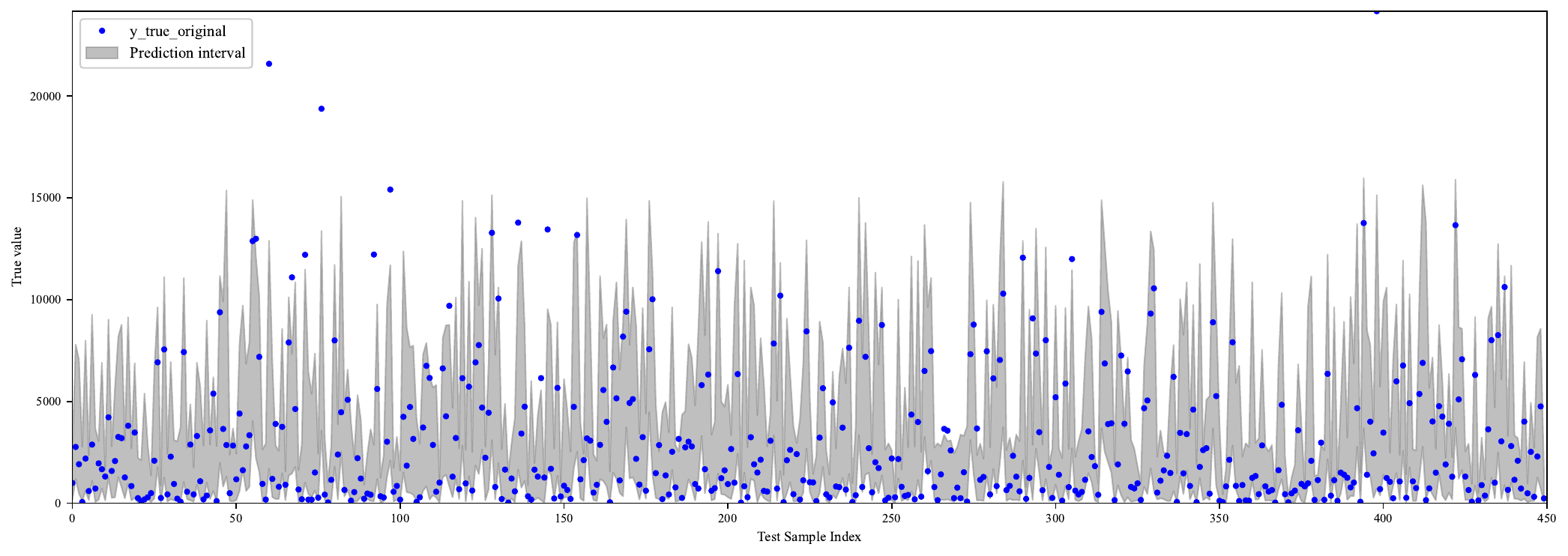}
    \caption{Interval prediction results versus true values. Blue dots are observations; grey shaded areas denote prediction intervals.}
    \label{fig:8}
\end{figure}

Figure~\ref{fig:7} shows the relationship between prediction errors and true values.  
Most predictions cluster near zero error, indicating high overall accuracy, while a few outliers exhibit larger deviations.  
Figure~\ref{fig:8} shows that most observed AADT values fall within the prediction intervals, confirming model reliability, particularly for mid- and low-AADT ranges.  
Higher variability in large-AADT values suggests potential gains from refined hyperparameter tuning or outlier treatment.

\section{Feature Groups}
This section provides a detailed summary of the feature groups used in the study. 
Table~\ref{tab:feature_groups} contains the complete list.

\begin{table}[H]
\caption{Feature Groups Summary}
\label{tab:feature_groups}
\centering
\footnotesize
\renewcommand{\arraystretch}{1.1}
\begin{tabular}{p{3.8cm} p{4cm} p{4.5cm} p{2.3cm}}
\toprule
\textbf{Group Name} & \textbf{Description} & \textbf{Variable Types} & \textbf{Buffer (m)} \\
\midrule
\multicolumn{4}{l}{\textbf{Basic Geographic Features (3 groups)}} \\
group\_latitude & Geographic latitude information & latitude & N/A \\
group\_longitude & Geographic longitude information & longitude & N/A \\
group\_Class10\_code & Classification codes for land use & Class10\_code & N/A \\
\midrule
\multicolumn{4}{l}{\textbf{Accessibility Features (1 group)}} \\
group\_accessibility & Accessibility measures to Built-Up Areas (BUA), Functional Urban Areas (FUA), and Target Cities (TCITY) 
& BUA\_bool, FUA\_bool \newline
  FUA\_Access (d1) \newline
  FUA\_Access (d2) \newline
  FUA\_Access (exp)\newline
  FUA\_edge\_dist \newline
  TCITY\_bool, TCITY\_Access (d1) \newline
  TCITY\_Access (d2) \newline
  TCITY\_Access (exp) & N/A \\
\midrule
\multicolumn{4}{l}{\textbf{Road Features (1 group)}} \\
group\_road & Road characteristics including length, class, function, type 
& Road\_length, Road\_class \newline
  Road\_function, Road\_primary \newline
  Road\_trunkRoad, Road\_formOfWay & N/A \\
\midrule
\multicolumn{4}{l}{\textbf{Transportation Infrastructure (2 groups)}} \\
group\_ports & Port infrastructure & All columns containing ``Port'' & N/A \\
group\_airports & Airport infrastructure & All columns containing ``Airport'' & N/A \\
\midrule
\multicolumn{4}{l}{\textbf{Spatial Lag Features (1 group)}} \\
group\_lag & Spatial lag variables capturing spillover effects & All columns containing ``lag'' & N/A \\
\midrule
\multicolumn{4}{l}{\textbf{Buffer Zone Features (7 groups across multiple radii)}} \\
group\_BCount\_[buffer] & Entity counts within buffer zones & BCount related columns & 500, 800, 1000, 1600, 2000, 3200 \\
group\_junc\_[buffer] & Junction counts within buffer zones & junc related columns & 500, 800, 1000, 1600, 2000, 3200 \\
group\_transport\_[buffer] & Railway, metro, bus stops within buffer zones & Railway, Metro, Bus & 500, 800, 1000, 1600, 2000, 3200 \\
group\_employment\_[buffer] & Employment within buffer zones & Emp related columns & 500, 800, 1000, 1600, 2000, 3200 \\
group\_population\_[buffer] & Population and households within buffer zones & Popu, HH\_total & 500, 800, 1000, 1600, 2000, 3200 \\
group\_vehicles\_[buffer] & Vehicle counts within buffer zones & Veh related columns & 500, 800, 1000, 1600, 2000, 3200 \\
group\_earnings\_[buffer] & Earnings within buffer zones & Earn related columns & 500, 800, 1000, 1600, 2000, 3200 \\
\midrule
\multicolumn{4}{l}{\textbf{Density Variables (1 group)}} \\
group\_density & MSOA-level employment and population density (gravity-based $\rho_G$ and exponential $\rho_E$, $\alpha=1.0,1.5,2.0$) & [metric]\_employment, [metric]\_population & N/A \\
\midrule
\multicolumn{4}{l}{\textbf{Summary}} \\
Total Feature Groups & \multicolumn{3}{c}{51} \\
Buffer Sizes (m) & \multicolumn{3}{c}{500, 800, 1000, 1600, 2000, 3200} \\
\bottomrule
\end{tabular}
\end{table}

\end{document}